\documentclass[letterpaper]{article} 
\usepackage{aaai23}  
\usepackage{times}  
\usepackage{helvet}  
\usepackage{courier}  
\usepackage[hyphens]{url}  
\usepackage{graphicx} 
\usepackage{natbib}  
\usepackage{caption} 
\usepackage{algorithmic}
\usepackage[linesnumbered,ruled,noend]{algorithm2e} 
\usepackage{amsmath,amssymb,amsfonts}
\usepackage{bm}
\usepackage{textcomp}
\usepackage{amsthm}
\usepackage{subfigure}
\usepackage{xcolor}
\usepackage{multirow}
\usepackage{booktabs}
\usepackage{siunitx}
\usepackage{cite}

\usepackage[title]{appendix}
\usepackage{newfloat}
\DeclareCaptionStyle{ruled}{labelfont=normalfont,labelsep=colon,strut=off} 

\title{Learnable Path in Neural Controlled Differential Equations}
\author{
    Sheo Yon Jhin\equalcontrib,\textsuperscript{\rm 1}
    Minju Jo\equalcontrib,\textsuperscript{\rm 1}
    Seungji Kook,\textsuperscript{\rm 1}
    Noseong Park,\textsuperscript{\rm 1}
    Sungpil Woo,\textsuperscript{\rm 2}
    Sunhwan Lim\textsuperscript{\rm 2}
}
\affiliations{
    \textsuperscript{\rm 1} Yonsei University, Seoul, South Korea,\\ \textsuperscript{\rm 2} Electronics and Telecommunications Research Institute (ETRI), Daejeon, South Korea\\
    \textsuperscript{\rm 1} {sheoyonj,alflsowl12,202132139,noseong}@yonsei.ac.kr,\\\textsuperscript{\rm 2}{woosungpil,shlim}@etri.re.kr
}

\begin{document}

\maketitle

\begin{abstract}
Neural controlled differential equations (NCDEs), which are \emph{continuous} analogues to recurrent neural networks (RNNs), are a specialized model in (irregular) time-series processing. In comparison with similar models, e.g., neural ordinary differential equations (NODEs), the key distinctive characteristics of NCDEs are i) the adoption of the continuous path created by an interpolation algorithm from each raw discrete time-series sample and ii) the adoption of the Riemann--Stieltjes integral. It is the continuous path which makes NCDEs be analogues to continuous RNNs. However, NCDEs use existing interpolation algorithms to create the path, which is unclear whether they can create an optimal path. To this end, we present a method to generate another latent path (rather than relying on existing interpolation algorithms), which is identical to learning an appropriate interpolation method. We design an encoder-decoder module based on NCDEs and NODEs, and a special training method for it. Our method shows the best performance in both time-series classification and forecasting.
\end{abstract}

\section{Introduction}

Deep learning for time-series data is one of the most active research fields in machine learning since many real-world applications deal with time-series data. For instance, time-series forecasting is a long-standing research problem, ranging from stock price forecasting to climate and traffic forecasting~\cite{reinsel2003elements,fu2011review,bing2018stgcn,wu2019graphwavenet,guo2019astgcn,song2020stsgcn,huang2020lsgcn,NEURIPS2020_ce1aad92,chen2021ZGCNET,fang2021STODE,choi2022STGNCDE,hwang2021climate}. Time-series classification ~\cite{fawaz2019deep} and time-series anomaly detection ~\cite{zhang2019deep} are also popular. Recurrent neural networks (RNNs), such as LSTM ~\cite{sepp1997long}, GRU ~\cite{chung2014empirical} and so on, have been typically used for these purposes. Recently, however, researchers extended the basis of the time-series processing model design to differential equations (far beyond classical RNNs), e.g., neural ordinary differential equations (NODEs) and neural controlled differential equations (NCDEs). It is already well known that much scientific time-series data can be clearly described by differential equations ~\cite{NIPS2018_7892,NEURIPS2019_42a6845a,NEURIPS2020_4a5876b4}. For instance, there exist various differential equation-based models for physical, social scientific, and financial phenomena and some of them received Nobel prizes, e.g., the Black–Scholes–Merton model~\cite{10.2307/1831029}. In this regard, we consider that differential equation-based approaches are one of the most suitable strategies in designing time-series models, especially for real-world data.
\begin{figure*}
    \centering
    \subfigure[The architecture of NCDE]{\includegraphics[width=0.6\columnwidth]{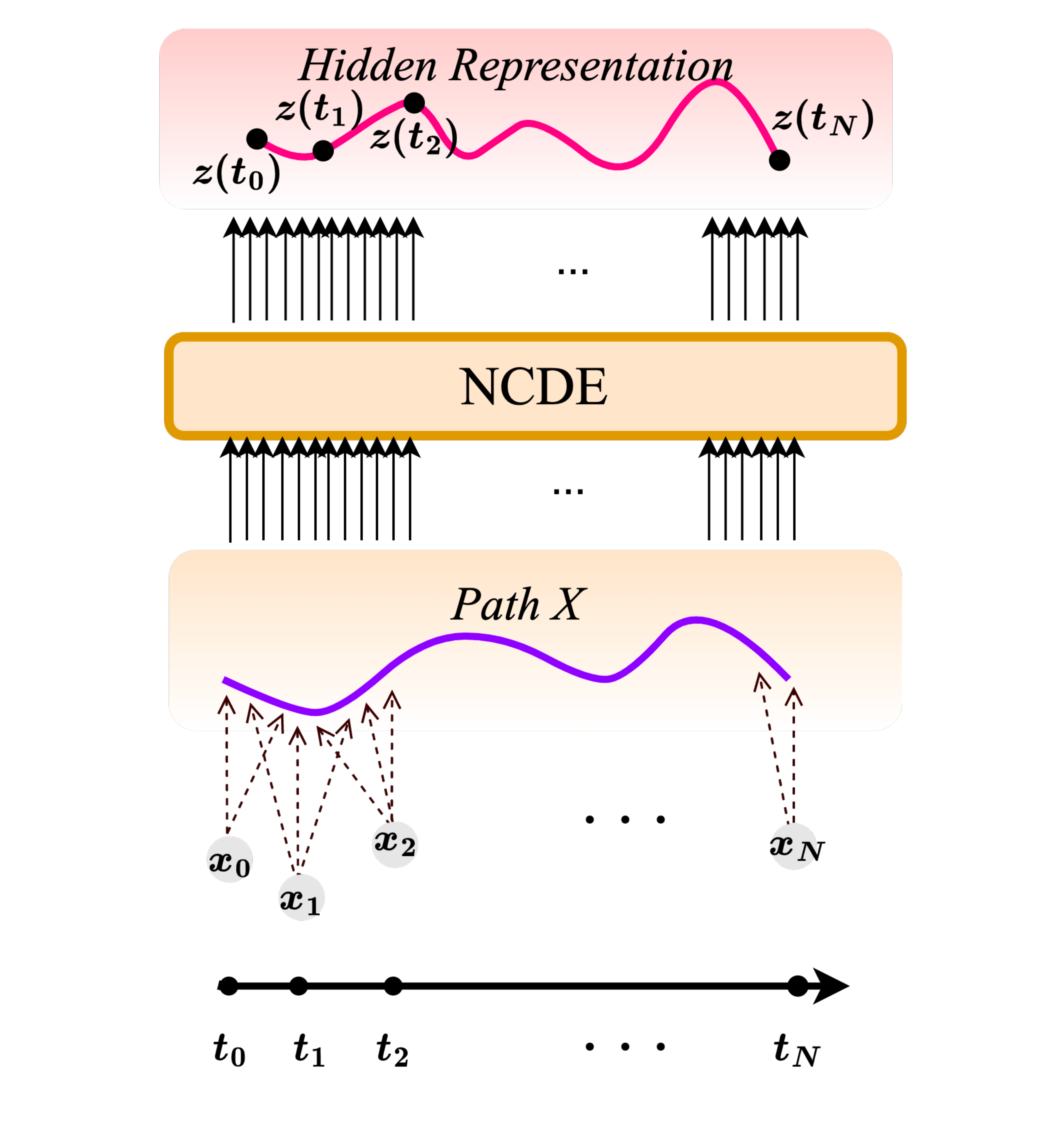}}
    \subfigure[The architecture of LEAP]{\includegraphics[width=1.0\columnwidth]{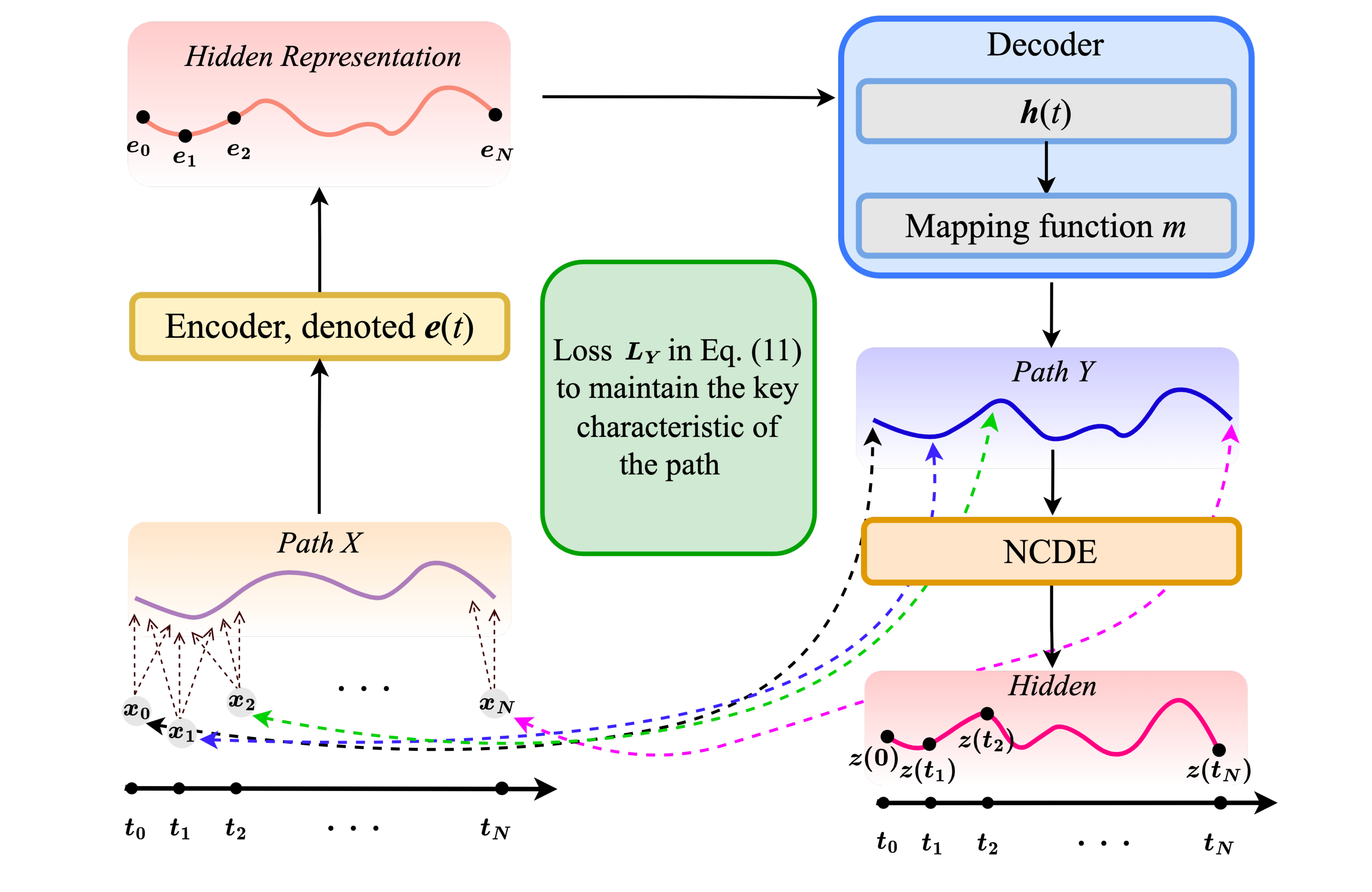}}
    \caption{(a) is the architecture of NCDE where the path $X$ is created by interpolating $\{(\mathbf{x}_i, t_i)\}_{i=0}^N$. (b) is the proposed architecture of LEAP. Our proposed concept corresponds to learning an interpolation method to generate the path $Y$.}
    \label{fig:archi_ncde_leap}
\end{figure*}
Neural controlled differential equations (NCDEs)~\cite{NEURIPS2020_4a5876b4} are breakthrough concepts to interpret recurrent neural networks (RNNs) in a continuous manner. While the initial value determines the solution of differential equation in neural ordinary differential equations (NODEs)~\cite{NIPS2018_7892}, NCDEs keep reading time-evolving values (as in RNNs) and their solutions are determined by the entire input. In this regard, NCDEs are a continuous analogue to RNNs and they show the state-of-the-art performance for many time-series datasets. We compare the solution $\mathbf{z}(T)$ of NODEs and NCDEs as follows:
\begin{enumerate}
    \item For NODEs, \begin{align}\label{eq:node}\mathbf{z}(T) = \mathbf{z}(0) + \int_{0}^{T} f(\mathbf{z}(t), t;\mathbf{\theta}_f) dt; \end{align}
    \item For NCDEs, \begin{align}\label{eq:ncde}\mathbf{z}(T) &= \mathbf{z}(0) + \int_{0}^{T} f(\mathbf{z}(t);\mathbf{\theta}_f) dX(t)\\&= \mathbf{z}(0) + \int_{0}^{T} f(\mathbf{z}(t);\mathbf{\theta}_f) \frac{dX(t)}{dt} dt,\label{eq:ncde2}\end{align} where $X(t)$ is a continuous path created from a raw discrete time-series sample $\{(\mathbf{x}_i, t_i)\}_{i=0}^N$ by an interpolation algorithm, where $t_i$ means the time-point of the observation $\mathbf{x}_i$, $t_0 = 0$, $t_N = T$ and $t_i < t_{i+1}$ (cf. Fig.~\ref{fig:archi_ncde_leap}(a)). Note that NCDEs keep reading the derivative of $X(t)$ over time, denoted $\dot{X}(t) \stackrel{\text{def}}{=} \frac{dX(t)}{dt}$, whereas NODEs do not. Given fixed $\mathbf{\theta}_f$ (after training), $\mathbf{z}(0)$ solely determines the evolutionary trajectory from $\mathbf{z}(0)$ to $\mathbf{z}(T)$ in NODEs. 
\end{enumerate}

In NCDEs, the time-derivative of $X(t)$, $\frac{dX(t)}{dt}$, is considered to define the time-derivative of $\mathbf{z}(t)$, $\frac{d\mathbf{z}(t)}{dt}$, which means they keep reading values from $\frac{dX(t)}{dt}$. Thus, selecting the appropriate interpolation method for creating the continuous path $X$ is crucial. The first work proposing the NCDE concept prefers the natural cubic spline algorithm for its suitable characteristic to be used in NCDEs, i.e., the path created by the natural cubic spline is twice differentiable and continuous. As reported in~\cite{morrill2021neural}, however, other interpolation algorithms can also be used. According to their experimental results, however, there does not exist a clear winning interpolation method that works always well although the natural cubic spline is, in general, a good choice.


To this end, we propose to learn an interpolation method optimized for a downstream task, i.e., \underline{\textbf{LEA}}rnable \underline{\textbf{P}}ath (LEAP). We insert an encoder-decoder module to generate a path $Y$, rather than relying on the path $X$ created by an interpolation algorithm (cf. Fig.~\ref{fig:archi_ncde_leap}(b)). The additional module reads $X$ to generate another fine-tuned path $Y$ that will be used to define the hidden vector $\mathbf{z}$ over time $t$. One can consider that our method learns how to interpolate and generate the path $Y$.


We conducted time-series classification and forecasting experiments with four datasets and twelve baselines, which are all well-recognized standard benchmark environments. Our method shows the best performance in terms of various accuracy and error metrics. Our contributions can be summarized as follows:
\begin{enumerate}
    \item We learn a path to be used to evolve $\mathbf{z}(t)$, which corresponds to learning an interpolation algorithm. 
    \item Our method outperforms existing baselines in all the cases in both time-series classification and forecasting.
\end{enumerate}

\section{Related Work and Preliminaries}

We introduce our literature survey for recent deep learning work related to time-series processing. We also deliver base knowledge to understand our paper.

\paragraph{Neural Ordinary Differential Equations} NODEs can process time-series data in a \emph{continuous} manner, which means they can read and write values at any arbitrary time-point $t$ by using the differential equation in Eq.~\eqref{eq:node}.

$\mathbf{z}(t) \in \mathbb{R}^D$, where $t \in [0,T]$, means a $D$-dimensional vector --- we use boldface to denote vectors and matrices. $\dot{\mathbf{z}}(t) \stackrel{\text{def}}{=} \frac{d\mathbf{z}(t)}{dt}$ is approximated by the neural network $f(\mathbf{z}(t), t;\mathbf{\theta}_f)$, and we need to solve the initial value problem to derive the final value $\mathbf{z}(T)$ from the initial value $\mathbf{z}(0)$, which is basically an integral problem. To solve the problem, we typically rely on existing ODE solvers, such as the explicit Euler Method, the 4th order Runge--Kutta (RK4) method, the Dormand--Prince (DOPRI) method, and so forth. In particular, the explicit Euler method, denoted $\mathbf{z}(t+h) = \mathbf{z}(t) + h f(\mathbf{z}(t), t;\mathbf{\theta}_f)$ where $h \in \mathbb{R}$ is a step-size parameter, is identical to the residual connection. Therefore, NODEs are considered as continuous analogues to residual networks.

There exist several popular time-series processing models based on NODEs, such as Latent-ODE, GRU-ODE, ACE-NODE, and so forth. Latent-ODE is an encoder-decoder architecture for processing time-series data. GRU-ODE showed that the GRU cell can be modeled by a differential equation. ACE-NODE is the first NODE-based model with an attention mechanism.

\paragraph{Neural Controlled Differential Equations} NCDEs in Eq.~\eqref{eq:ncde} are technically more complicated than NODEs. NCDEs use the Riemann--Stieltjes integral, as shown in Eq.~\eqref{eq:ncde2}, whereas NODEs use the Riemann integral. To solve Eq.~\eqref{eq:ncde2}, existing ODE solvers can also be used since $\dot{\mathbf{z}}(t) \stackrel{\text{def}}{=} \frac{d\mathbf{z}(t)}{dt} = f(\mathbf{z}(t);\mathbf{\theta}_f) \frac{dX(t)}{dt}$ in NCDEs. Regardless of the integral problem type, existing ODE solvers can be used once $\dot{\mathbf{z}}(t)$ can be properly modeled and calculated. Additionally, NCDEs are considered as a \emph{continuous} analogue to RNNs since they continuously read values $\dot{X}(t)$ over time. 

In order to use NCDEs, however, we need to create a continuous path $X$ from each raw discrete time-series sample, for which we typically use an interpolation algorithm. Among various methods, the natural cubic spline method is frequently used. As reported in~\cite{morrill2021neural}, the model accuracy of NCDEs is greatly influenced by how to interpolate discrete time-series data. To this end, we propose to let a neural network interpolate rather than relying on existing interpolation algorithms.

In ~\cite{jhin2021ancde}, they proposed ANCDEs to insert an attention mechanism into NCDEs. Unlike our method, however, their goal is to pick useful information from the path $X$. Therefore, their method does not learn a new interpolation method whereas our proposed concept is equivalent to learning an interpolation method for a downstream task.

\section{Proposed Method}

We describe our proposed method that is to learn a path, which corresponds to learning an interpolation algorithm (rather than relying on existing algorithms). In other words, our proposed model generates another latent path $Y$ from the path $X$ for a downstream task.

\paragraph{Overall Design}


Fig.~\ref{fig:archi_ncde_leap}(b) shows the detailed design of our method, LEAP. The overall workflow is as follows:
\begin{enumerate}
    \item The path $X$ is created from a discrete time-series sample by an existing interpolation algorithm.
    \item The encoder NCDE, the yellow box in Fig.~\ref{fig:archi_ncde_leap}(b), produces a hidden vector at time $t$, denoted $\mathbf{e}(t)$.
    \item We use the hidden vector at $t_N$, i.e, $\mathbf{e}(t_N)$, to generate the hidden representation of the input time-series.
    \item From the hidden representation, the NODE-based decoder, the blue box in Fig.~\ref{fig:archi_ncde_leap}(b), produces another latent path $Y$.
    \item There is a loss to maintain the key characteristics of the path $X$ while generating the path $Y$ (cf. Eq.~\eqref{eq:loss}). Therefore, the path $Y$ is not simply a latent path but a latent path fine-tuned from $X$.
    \item After that, there is one more NCDE based on $Y$ for a downstream task, the orange box in Fig.~\ref{fig:archi_ncde_leap}(b).
\end{enumerate}

We describe each part in detail, followed by a theoretical result that training the proposed model is well-posed.

\paragraph{Encoder-Decoder Module}

We first introduce our formulation to define the proposed LEAP. The entire module can be written, when we adopt the proposed encoder-decoder strategy to learn another path $Y$, as follows:
\begin{align}
    \mathbf{z}(T) &= \mathbf{z}(0) + \int_{0}^{T} g(\mathbf{z}(t);\mathbf{\theta}_g) \frac{dY(t)}{dt} dt,\label{eq:top}\\
    Y(t) &= m(\mathbf{h}(t);\mathbf{\theta}_m),\label{eq:map}\\
    \mathbf{h}(T) &= \mathbf{h}(0) + \int_{0}^{T} f(\mathbf{h}(t), t;\mathbf{\theta}_f) dt,\label{eq:mid}\\
    \mathbf{e}(T) &= \mathbf{e}(0) + \int_{0}^{T} k(\mathbf{e}(t);\mathbf{\theta}_k) \frac{dX(t)}{dt} dt,\label{eq:bottom}
\end{align}where $\mathbf{h}(0) = \phi_\mathbf{h}(\mathbf{e}(t_N); \mathbf{\theta}_{\phi_\mathbf{h}})$ for a raw discrete time-series sample $\{(\mathbf{x}_i, t_i)\}_{i=0}^N$, $Y$ is a path created from $X$, $\mathbf{z}$ is controlled by $Y$, and $\phi_\mathbf{h}$ is a fully-connected layer-based transformation. As mentioned earlier, $X$ is created by the natural cubic spline algorithm from raw discrete time-series observations. Even though the natural cubic spline algorithm is able to produce a path suitable for NCDEs, it is hard to say that the algorithm is always the best option~\cite{morrill2021neural}. Thus, we propose to learn another path $Y$ which is optimized for a downstream machine learning task.  Then, $Y$ can be considered as a fine-tuned path from $X$ for a downstream machine learning task. We also note that $\dim(Y(t)) = \dim(X(t))$. Since $\dot{Y}(t)\stackrel{\text{def}}{=}\frac{dY(t)}{dt} = \frac{dY(t)}{d\mathbf{h}(t)} \frac{d\mathbf{h}(t)}{dt}$ by the chain rule, Eq.~\eqref{eq:top} can be rewritten as follows:
\begin{align}\label{eq:lp}
\mathbf{z}(T) &= \mathbf{z}(0) + \int_{0}^{T} g(\mathbf{z}(t);\mathbf{\theta}_g) \frac{dY(t)}{d\mathbf{h}(t)} f(\mathbf{h}(t), t;\mathbf{\theta}_f) dt,
\end{align}where $\frac{dY(t)}{d\mathbf{h}(t)}$ is defined by the mapping function $m$ and can be easily calculated by the automatic differentiation method. For instance, $\frac{dY(t)}{d\mathbf{h}(t)}$ will be simply $\mathbf{W}$ if $m(\mathbf{h}(t);\mathbf{\theta}_m) = \mathbf{W}\mathbf{h}(t) $, i.e., $m$ is a zero-biased fully connected layer where $\mathbf{\theta}_m = \mathbf{W}$.

Therefore, our proposed concept can be rather succinctly described by Eq.~\eqref{eq:lp}. However, the key part in our method lies in training $\mathbf{\theta}_f$ and $\mathbf{\theta}_m$ to generate $\mathbf{h}(t)$ and $Y(t)$, i.e, the decoder part. We want that $Y$ i) shares important characteristics with $X$ to ensure the theoretical correctness of the proposed method and ii) at the same time produces a better path for a downstream machine learning task than $X$. To this end, we define the following augmented NCDE:
\begin{align}
\frac{d}{dt}{\begin{bmatrix}
  \mathbf{z}(t) \\
  \mathbf{h}(t) \\
  \end{bmatrix}\!} = {\begin{bmatrix}
  g(\mathbf{z}(t);\mathbf{\theta}_g) \frac{d\mathbf{Y}(t)}{d\mathbf{h}(t)} f(\mathbf{h}(t), t;\mathbf{\theta}_f) \\
  f(\mathbf{h}(t), t;\mathbf{\theta}_f)\\
  \end{bmatrix}\!},
\end{align}where the initial values are defined as follows:
\begin{align}
{\begin{bmatrix}
  \mathbf{z}(0) \\
  \mathbf{h}(0) \\
  \end{bmatrix}\!} = {\begin{bmatrix}
  \phi_{\mathbf{z}}(X(0));\mathbf{\theta}_{\phi_{\mathbf{z}}}) \\
  \phi_\mathbf{h}(\mathbf{e}(t_N));\mathbf{\theta}_{\phi_\mathbf{h}}) \\
  \end{bmatrix}\!},
\end{align}where $\phi_{\mathbf{z}}$ and $\phi_\mathbf{h}$ are fully-connected layer-based trainable transformation functions to generate the initial values.

In addition, let $\mathbf{z}(T)$ be the last hidden vector. We have an output layer with a typical construction based on $\mathbf{z}(T)$. The output layer is same as that in the original NCDE model. For forecasting (regression), we use a fully-connected layer whose output size is the same as the size of prediction.

\paragraph{How To Train}

\begin{algorithm}[t]
\small
\SetAlgoLined
\caption{How to train LEAP}\label{alg:train}
\KwIn{Training data $D_{train}$, Validating data $D_{val}$, Maximum iteration numbers $max\_iter$}
Initialize $\mathbf{\theta}_f$, $\mathbf{\theta}_g$, $\mathbf{\theta}_k$, $\mathbf{\theta}_m$, and other parameters, denoted $\mathbf{\theta}_{others}$, if any, e.g., the parameters of the output layer;

$i \gets 0$;

\While {$i < max\_iter$}{
    Train $\mathbf{\theta}_f$, $\mathbf{\theta}_g$, $\mathbf{\theta}_k$, $\mathbf{\theta}_m$, and $\mathbf{\theta}_{others}$ using the loss $L$\;\label{alg:train3}
    
    Validate and update the best parameters, $\mathbf{\theta}^*_f$, $\mathbf{\theta}^*_g$, $\mathbf{\theta}^*_k$, $\mathbf{\theta}^*_m$, and $\mathbf{\theta}^*_{others}$, with $D_{val}$\;
    
    $i \gets i + 1$;
}
\Return $\mathbf{\theta}^*_f$, $\mathbf{\theta}^*_g$, $\mathbf{\theta}^*_k$, $\mathbf{\theta}^*_m$, and $\mathbf{\theta}^*_{others}$;
\end{algorithm}

We use the following MSE and log-density loss definitions given training data $D_{train}$ to define $L_Y$:
\begin{align}\begin{split}\label{eq:loss}
    L_{Y} \stackrel{\text{def}}{=}&  \frac{\sum_{j=1}^{M}\sum_{i=1}^{N}\alpha\|Y(t^{(j)}_i) - \mathbf{x}^{(j)}_i\|_2^2 - \beta \log p(\hat{\mathbf{h}}(t^{(j)}_i))}{MN},
\end{split}\end{align}where $M = |D_{train}|$ is the size of training data. The final loss $L$ is the sum of $L_{Y}$ and a task specific loss $L_{task}$, e.g., the cross-entropy loss for time-series classification or the mean squared error (MSE) loss for time-series forecasting. $\alpha$ and $\beta$ are the coefficients of the two terms in Eq.~\eqref{eq:loss}. Algorithm~\ref{alg:train} shows our training algorithm.

Given the $j$-th time-series sample in our training data, denoted $\{(\mathbf{x}^{(j)}_i, t^{(j)}_i)\}_{i=0}^N$, $Y(t^{(j)}_i) = \mathbf{x}^{(j)}_i$ is preferred as in the natural cubic spline. We inject this by using the MSE loss term of $L_Y$ in Eq.~\eqref{eq:loss}. In addition, we adopt the Hutchinson's unbiased estimator\footnote{Since NODEs are continuous and bijective, i.e., invertible, the change of variable theorem teaches us how to calculate the log-density --- as a matter of fact, many other invertible neural networks, such as Glow, RealNVP, ans so on, rely on the same theory for their explicit likelihood training. However, this requires non-trivial computation. The Hutchinson's statistical method can reduce the complexity of measuring the log-density. On the top of that, better way to measure the log-density for NODEs than the Hutchinson's estimator is not known for now~\cite{grathwohl2019ffjord}.} to measure the log-density $\log p(\hat{\mathbf{h}}(t^{(j)}_i))$, where $Y(t^{(j)}_i) = m(\hat{\mathbf{h}}(t^{(j)}_i);\mathbf{\theta}_m)$ and $\hat{\mathbf{h}}(t^{(j)}_i) = \mathbf{x}^{(j)}_{i-1} + \int_{t^{(j)}_{i-1}}^{t^{(j)}_i} f(\mathbf{h}(t); \mathbf{\theta}_f) dt$ as in Eqs.~\eqref{eq:map} and~\eqref{eq:mid}. We use $\hat{\mathbf{h}}$, which is created directly from the real observation $\mathbf{x}^{(j)}_{i-1}$, to distinguish it from $\mathbf{h}$. The Hutchinson's estimator can be defined as follows in our case --- refer to Appendix for the detailed explanation on the Hutchinson's estimator~\cite{jhinjo2023leap}:
\begin{align}\label{eq:hut}
    \log p(\hat{\mathbf{h}}(t^{(j)}_i)) &= \log p(\mathbf{x}^{(j)}_{i-1}) - \mathbb{E}_{p(\mathbf{\epsilon})}\Big[\int_{t_{i-1}^{(j)}}^{t_{i}^{(j)}} \mathbf{\epsilon}^\intercal\frac{\partial f}{\partial \mathbf{h}(t)}\mathbf{\epsilon} dt\Big],
\end{align}where $p(\mathbf{\epsilon})$ is a standard Gaussian or Rademacher distribution~\cite{doi:10.1080/03610919008812866}.  The time complexity to calculate the Hutchinson's estimator is slightly larger than that of evaluating $f$ since the vector-Jacobian product $\mathbf{\epsilon}^\intercal\frac{\partial f}{\partial \mathbf{h}(t)}$ has the same cost as that of evaluating $f$ using the reverse-mode automatic differentiation. Since $\log p(\mathbf{x}^{(j)}_{i-1})$ is a constant, minimizing the negative log-density is the same as minimizing $\mathbb{E}_{p(\mathbf{\epsilon})}\Big[\int_{t_{i-1}^{(j)}}^{t_{i}^{(j)}} \mathbf{\epsilon}^\intercal\frac{\partial f}{\partial \mathbf{h}(t)}\mathbf{\epsilon} dt\Big]$ only.

In order to perform the density estimation via the change of variable theorem, we need invertible layers. Unfortunately, it is not theoretically guaranteed that NCDEs are invertible. However, NODEs are always invertible (more specifically, homeomorphic). Therefore, we do not perform the density estimation with NCDEs.


\paragraph{Rationale Behind Our Loss Definition} For linear regression, it is known that the MSE training can achieve the maximum likelihood estimator (MLE) of model parameters. In general, however, the MSE training does not exactly return the MLE of parameters. Therefore, we mix the MLE training and the explicit likelihood training to yield the best outcome. The role of the MSE loss is to make $Y(t^{(j)}_i)$ and $\mathbf{x}^{(j)}_i$ as close as possible and then, the negative log-density training enhances its likelihood since $Y(t^{(j)}_i)$ is generated from $\hat{\mathbf{h}}(t^{(j)}_i)$.

\paragraph{Well-posedness} The well-posedness\footnote{A well-posed problem means i) its solution uniquely exists, and ii) its solution continuously changes as input data changes.} of NODEs and NCDEs was already proved in \cite[Theorem 1.3]{lyons2004differential} under the mild condition of the Lipschitz continuity. We show that our NODE/NCDE layers are also well-posed problems. Almost all activations, such as ReLU, Tanh, and Sigmoid, have a Lipschitz constant of 1. Other common neural network layers, such as dropout, batch normalization and other pooling methods, have explicit Lipschitz constant values.~\cite{NIPS2018_7892} Therefore, the Lipschitz continuity of $f$, $g$, $k$ can be fulfilled in our case. This makes our training problem well-posed. As a result, our training algorithm solves a well-posed problem so its training process is stable in practice.

\section{Experiments}\label{sec:exps}
In this section, we describe our experimental environments and results. We conduct experiments with time-series classification and forecasting. Our software and hardware environments are as follows: \textsc{Ubuntu} 18.04 LTS, \textsc{Python} 3.7.6, \textsc{Numpy} 1.20.3, \textsc{Scipy} 1.7, \textsc{Matplotlib} 3.3.1,  \textsc{CUDA} 11.0, and \textsc{NVIDIA} Driver 417.22, i9 CPU, and \textsc{NVIDIA RTX Titan}. We repeat the training and testing procedures with five different random seeds and report their mean and standard deviation of evaluation metrics.

\subsection{Experimental Environments}
\paragraph{Hyperparameters}
We list all the hyperparameter settings in Appendix.
\paragraph{Baselines}
For time series forecasting and classification experiments, we compare our method with the following methods. 
\begin{enumerate}
    \item RNN, LSTM~\cite{sepp1997long}, and GRU~\cite{chung2014empirical} are all recurrent neural network-based models that can process sequential data. LSTM is designed to learn long-term dependencies and GRU uses gating mechanisms to control the flow of information. 
    \item GRU-$\Delta t$ is a GRU that additionally takes as input the time difference between observations.
    \item GRU-D~\cite{che2016recurrent} is a modified version of GRU-$\Delta t$ with learnable exponential decay between observations.
    \item GRU-ODE~\cite{debrouwer2019gruodebayes,jordan2019gated} is a NODE similar to GRU. This model is a continuous counterpart of GRU.
    \item ODE-RNN~\cite{NEURIPS2019_42a6845a} is an extension of GRU-$\Delta t$ to NODE. 
    \item Latent-ODE~\cite{NEURIPS2019_42a6845a} is a suitable model for time-series in which the latent state follows NODE. In this work, we use the recognition network of the existing Latent-ODE model as ODE-RNN.
    \item Augmented-ODE is to augment the ODE state of Latent-ODE with the method proposed in~\cite{NIPS2019_8577}.
    \item ACE-NODE~\cite{jhin2021acenode} is the state-of-the-art attention-based NODE model which has dual co-evolving NODEs. 
    \item NCDE~\cite{NEURIPS2020_4a5876b4} is solved using controlled differential equations, which are well-understood mathematics.
    \item ANCDE~\cite{jhin2021ancde} is to insert an attention mechanism into NCDEs.
\end{enumerate}

\subsection{Time Series Classification Experimental Results}
We introduce our experimental results for time-series classification with the following three datasets. Character trajectories, Speech Commands, and PhysioNet Sepsis are the datasets collected from real-world applications. Evaluating the performance of our model with these datasets proves the competence of our model in various fields. We use the accuracy for balanced datasets and AUROC for imbalanced datasets.~\cite{NEURIPS2019_42a6845a,NEURIPS2020_4a5876b4}.

\paragraph{Character Trajectories}
The Character Trajectories dataset from the UEA time-series classification archive~\cite{bagnall2018uea} consists of values on the x-axis, y-axis and pen tip force of Latin alphabets as features. This dataset was collected while using a tablet with a sampling frequency of 200Hz, and there are 2,858 time-series character samples in total. The length of each data sample was truncated to 182 and each data has 3 dimensional vectors, i.e., x, y and pen tip force, and the length of each vector is 182. We used these features to classify alphabets into 20 classes (`a’, `b', `c', `d', `e', `g', `h', `l', `m', `n', `o', `p', `q', `r', `s', `u', `v', `w', `y', `z'), while other 6 letters were excluded from this task.

\begin{table}[t]
\scriptsize
\setlength{\tabcolsep}{1pt}
\centering

\begin{tabular}{ccccc} 
\hline
\multirow{2}{*}{Model} & \multicolumn{3}{c}{Test Accuracy} & \multirow{2}{*}{\begin{tabular}[c]{@{}c@{}}Memory\\Usage (MB)   \end{tabular}} \\ \cline{2-4}& 30\% dropped & 50\% dropped & 70\% dropped &                                                                           \\  \hline

GRU-$\Delta t$      & 0.941 ± 0.018     & 0.928 ± 0.021     & 0.918 ± 0.017     & 16.5    \\
GRU-D               & 0.938 ± 0.020     & 0.910 ± 0.048     & 0.916 ± 0.021     & 17.8    \\
GRU-ODE             & 0.894 ± 0.012     & 0.886 ± 0.039     & 0.891 ± 0.029     & 1.51    \\
ODE-RNN             & 0.946 ± 0.007     & 0.954 ± 0.003     & 0.949 ± 0.004     & 15.5    \\
Latent-ODE          & 0.881 ± 0.021     & 0.878 ± 0.021     & 0.879 ± 0.048     & 181     \\
Augmented-ODE       & 0.972 ± 0.012     & 0.948 ± 0.022     & 0.929 ± 0.023     & 186     \\
ACE-NODE            & 0.881 ± 0.036     & 0.879 ± 0.025     & 0.913 ± 0.028     & 113     \\
NCDE                & 0.988 ± 0.004     & 0.988 ± 0.002     & 0.985 ± 0.005     & 1.38    \\
ANCDE               & 0.988 ± 0.001     & 0.986 ± 0.002     & 0.986 ± 0.004     & 2.02    \\ \hline
\textbf{LEAP} & \textbf{0.992 ± 0.001} & \textbf{0.993 ± 0.002} & \textbf{0.991 ± 0.003}    & 9.24 \\ \hline
\end{tabular}
\caption{Accuracy (mean ± std, computed across five runs) on Irregular Character Trajectories}\label{tbl:charactertrajectory1}
\end{table}

\paragraph{Speech Commands}
The Speech Commands dataset is a one-second long audio data recorded with voice words such as ‘left’, ‘right’, ‘cat’, and ‘dog’ and noise heard in the background \cite{DBLP:journals/corr/abs-1804-03209}.
We use ‘yes’, ‘no’, ‘up’, ‘down’, ‘left’, ‘right’, ‘on’, ‘off’, ‘stop’ to solve a balanced classification problem among all 35 labels, using a total of 34975 time-series samples. The length (resp. the dimensionality) of each time-series is 161 (resp. 20).
\begin{table}[t]
\scriptsize
\centering
\begin{tabular}{ccc}
\hline
Model & Test Accuracy &Memory Usage (MB)\\ 
\hline
RNN             &  0.197 ± 0.006             &1,905     \\
LSTM            &  0.684 ± 0.034             &4,080     \\
GRU             &  0.747 ± 0.050             &4,609     \\
GRU-$\Delta t$  &  0.453 ± 0.313             &1,612     \\
GRU-D           &  0.346 ± 0.286             &1,717     \\
GRU-ODE         &  0.487 ± 0.018             &171.3     \\
ODE-RNN         &  0.678 ± 0.276             &1,472     \\
Latent-ODE      &  0.912 ± 0.006             &2,668     \\
Augmented-ODE   &  0.911 ± 0.008             &2,626     \\
ACE-NODE        &  0.911 ± 0.003             &3,046     \\
NCDE            &  0.898 ± 0.025             &174.9     \\
ANCDE           &  0.807 ± 0.075             &179.8     \\
\hline  
\textbf{LEAP}   & \textbf{0.922 ± 0.002}     &391.1     \\
\hline
\end{tabular}
\caption{Accuracy on Speech Commands}\label{tbl:speech}
\end{table}
\paragraph{PhysioNet Sepsis}
Since Sepsis~\cite{9005736,article} is a life-threatening illness leading many patients to death, early and correct diagnosis is important, which makes experiments on this dataset more meaningful. The PhysioNet 2019 challenge on sepsis prediction is originally a partially-observed dataset so that it fits to our irregular time-series classification experiments. Status of patients in ICU --- both static and time-dependent features --- is recorded in the dataset, and we only used 34 time-dependent features for our time-series classification. The goal for this classification is to predict the development of sepsis, which makes experiment binary classification. The dataset consists of 40,355 cases with variable time-series length, and about 90\% of data is missing. Due to the data imbalance with only a sepsis positive rate of 5\%, we evaluate our experiment using AUROC.

\begin{table}[t]
\scriptsize
\setlength{\tabcolsep}{1pt}
\centering

\begin{tabular}{ccc} 
\hline
Model & Test AUROC &Memory Usage (MB)\\  \hline
GRU-$\Delta t$  & 0.861 ± 0.002                 & 837    \\
GRU-D           & 0.878 ± 0.019                 & 889    \\
GRU-ODE         & 0.852 ± 0.009                 & 454    \\
ODE-RNN         & 0.874 ± 0.012                 & 696    \\
Latent-ODE      & 0.782 ± 0.014                 & 133    \\
Augmented-ODE   & 0.842 ± 0.017                 & 998    \\
ACE-NODE        & 0.823 ± 0.012                 & 194    \\
NCDE            & 0.872 ± 0.003                 & 244    \\
ANCDE           & 0.886 ± 0.002                 & 285    \\ \hline
\textbf{LEAP}   & \textbf{0.908 ± 0.004}        & 306    \\ \hline
\end{tabular}
\caption{AUROC on PhysioNet Sepsis}\label{tbl:PhysioNet}
\end{table}

\paragraph{Experimental Results}
Table~\ref{tbl:charactertrajectory1} summarizes the accuracy of Character Trajectories. In order to create challenging situations, we randomly selected 30\%, 50\%, and 70\% of the values from each sequence. Therefore, this is basically an irregular time-series classification, and many baselines show reasonable scores. The three GRU-based models are specialized in processing irregular time-series and outperform some other ODE-based models. However, CDE-based models, including LEAP, show the highest scores. Among them, LEAP is clearly the best. Our method maintains an accuracy larger than 0.99 across all the dropping settings.

For the Speech Commands dataset, we summarize the results in Table~\ref{tbl:speech}. As summarized, all RNN/LSTM/GRU-based models are inferior to other differential equation-based models. We consider that this is because of the dataset characteristic. This dataset contains many audio signal samples and it is obvious that those physical phenomena can be well modeled as differential equations. Among many differential equation-based models, the two NCDE-based models, NCDE and LEAP, show the reasonable performance. However, LEAP significantly outperforms all others including NCDE. One more point is that our method requires much smaller GPU memory in comparison with many other baselines.

The time-series classification with PhysioNet Sepsis in Table~\ref{tbl:PhysioNet} is one of the most widely used benchmark experiments. Our method, LEAP, shows the best AUROC and its GPU memory requirement is smaller than many other baselines. For this dataset, all CDE-based models show the highest performances.

\subsection{Time Series Forecasting Experimental Results}
We introduce our experimental results for time-series forecasting with the following dataset. We pick 1 dataset, MuJoCo, to evaluate the model's forecasting performance. Since MuJoCo is physics engine simulation data, using this data to predict future trajectories is an important study for mechanics and natural sciences. In addition, we use the MSE for our main metric, which is a metric generally used in time-series forecasting~\cite{NEURIPS2019_42a6845a,debrouwer2019gruodebayes,NEURIPS2020_4a5876b4}.
\paragraph{MuJoCo} This dataset was generated from 10,000 simulations of the ``Hopper'' model using the DeepMind Control Suite. This physics engine supports research in robotics, machine learning, and other fields that require accurate simulation, such as biomechanics. The dataset is 14-dimensional, and 10,000 sequences of 100 regularly-sampled time points each \cite{DBLP:journals/corr/abs-1801-00690}. The default training/testing horizon in MuJoCo is reading the first 50 observations in a sequence and forecasting the last 10 observations.

\begin{table}[t]
\scriptsize
\setlength{\tabcolsep}{1pt}
\centering

\begin{tabular}{ccccc} 
\hline
\multirow{2}{*}{Model} & \multicolumn{3}{c}{Test MSE} & \multirow{2}{*}{\begin{tabular}[c]{@{}c@{}}Memory\\Usage (MB)   \end{tabular}} \\ \cline{2-4}& 30\% dropped & 50\% dropped & 70\% dropped &                                                             \\  \hline
GRU-$\Delta t$  & 0.186 ± 0.036     & 0.189 ± 0.015     & 0.187 ± 0.018                     & 533   \\
GRU-D           & 0.417 ± 0.032     & 0.421 ± 0.039     & 0.438 ± 0.042                     & 569   \\
GRU-ODE         & 0.826 ± 0.015     & 0.821 ± 0.015     & 0.681 ± 0.014                     & 146   \\
ODE-RNN         & 0.242 ± 0.213     & 0.240 ± 0.110     & 0.240 ± 0.116                     & 115   \\
Latent-ODE      & 0.048 ± 0.001     & 0.043 ± 0.004     & 0.056 ± 0.001                     & 314   \\
Augmented-ODE   & 0.042 ± 0.004     & 0.048 ± 0.005     & 0.052 ± 0.003                     & 286   \\
ACE-NODE        & 0.047 ± 0.007     & 0.047 ± 0.005     & 0.048 ± 0.005                     & 423   \\
NCDE            & 0.028 ± 0.000     & 0.029 ± 0.001     & 0.031 ± 0.004                     & 52.1  \\
ANCDE           & 0.035 ± 0.002     & 0.031 ± 0.003     & 0.033 ± 0.003                     & 79.2  \\\hline
\textbf{LEAP}   & \textbf{0.022  ± 0.001} & \textbf{0.022 ± 0.002} & \textbf{0.022 ± 0.001} & 144   \\ \hline

\end{tabular}

\caption{MSE on Irregular MuJoCo}\label{tbl:mujoco}
\end{table}

\paragraph{Experimental Results}
We drop random 30\%, 50\%, and 70\% values, i.e., irregular time-series forecasting, to create challenging environments. In Table~\ref{tbl:mujoco}, our method, LEAP, clearly shows the best MSE for all dropping ratios. One outstanding point in our model is that the MSE is not greatly influenced by the dropping ratio but maintains its small error across all the dropping ratios.

\subsection{Ablation and Sensitivity Studies} \label{sec:ablation}
\paragraph{Sensitivity to $\alpha, \beta$ with Fixed Ratio 1.}
\begin{figure}[!t]
    \centering
    \subfigure[Sensitivity to $\alpha, \beta$ in Character Trajectories]{\includegraphics[width=0.49\columnwidth]{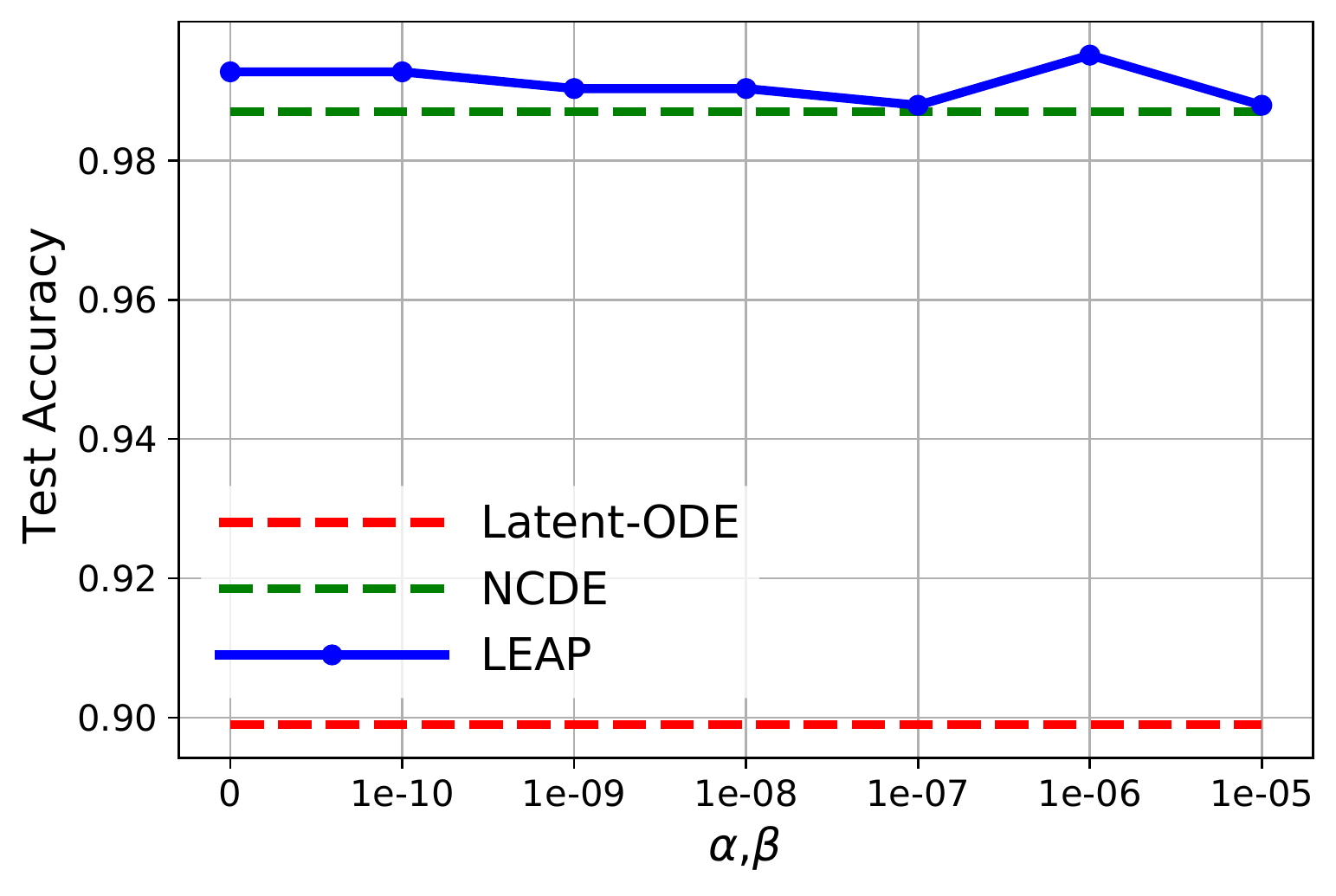}} \hfill
    \subfigure[Error by the output length in MuJoCo]{\includegraphics[width=0.49\columnwidth,trim={0 0.1cm 0 0},clip]{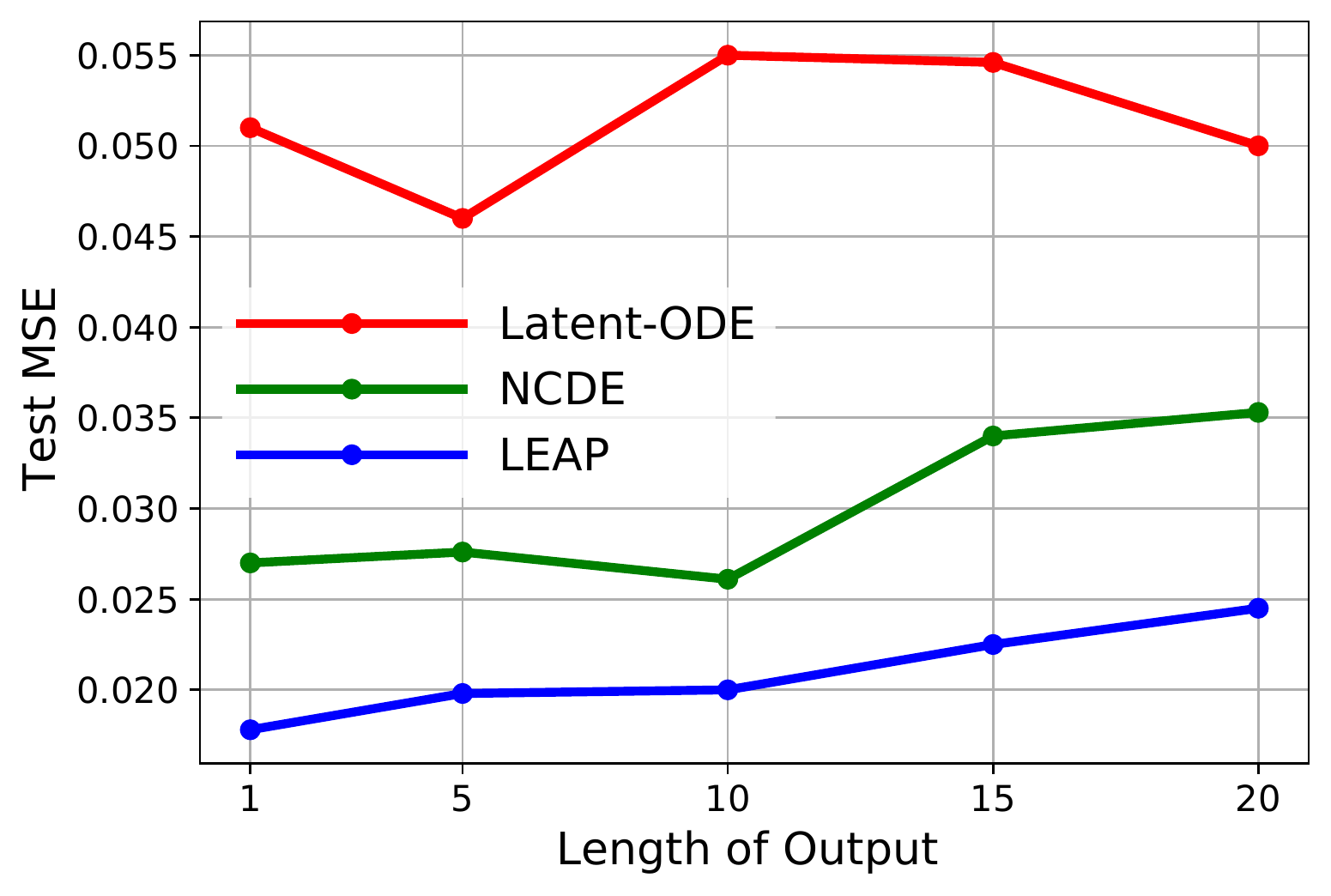}}
    \caption{Two sensitivity and ablation analyses. More figures are in Appendix.} 
    
    \label{fig:ratio_fix_outputseq}
\end{figure}

Fig.~\ref{fig:ratio_fix_outputseq}(a) shows the sensitivity curve w.r.t. $\alpha, \beta$, while setting $\alpha$ and $\beta$ to same value, i.e. the ratio of $\alpha$ to $\beta$ is 1, as in our original setting. 
With all the $\alpha$, $\beta$ settings, LEAP always shows better accuracy than those of the baselines, which shows the efficacy of our model. We also conduct experiments for the sensitivity to $\alpha, \beta$ with various ratios, and these results are in Appendix for space reasons.


\paragraph{Ablation on the Output Sequence Length}
We also compare our model with NCDE and Latent-ODE by varying the length of output (forecasting). After fixing the input length to 50, we variate the output length in $\{1, 5, 10, 15, 20\}$. As shown in Fig.~\ref{fig:ratio_fix_outputseq}(b), our proposed method consistently outperforms others. Moreover, MSE of LEAP's predicting 20-length sequence is lower than that of other models' predicting 1-length sequence. These results show that new latent paths made by LEAP skillfully represent the whole stream of data and capture parts which should be emphasized regardless of output sequence length.

\begin{table}[t]
\scriptsize
\setlength{\tabcolsep}{4pt}
\centering

\begin{tabular}{ccccc}
\hline
\multirow{3}{*}{Forecasting Horizon} & \multicolumn{4}{c}{Model being compared to LEAP}                                                                  \\ \cline{2-5}
                      & \multicolumn{2}{c}{GRU-$\Delta t$}                    & \multicolumn{2}{c}{NCDE}          \\ \cline{2-5} 
                      & \multicolumn{1}{c}{t-score} & \multicolumn{1}{c}{p-value} & \multicolumn{1}{c}{t-score} & p-value \\ \hline
1                     & \multicolumn{1}{c}{-88.95}   & \multicolumn{1}{c}{0.0}        & \multicolumn{1}{c}{-28.64}   & 1.41$\times e^{-114}$        \\ 
2                     & \multicolumn{1}{c}{-88.34}   & \multicolumn{1}{c}{0.0}        & \multicolumn{1}{c}{-29.85}   & 6.09$\times e^{-121}$         \\
3                     & \multicolumn{1}{c}{-87.19}   & \multicolumn{1}{c}{0.0}        & \multicolumn{1}{c}{-30.14}   & 1.74$\times e^{-122}$         \\
4                     & \multicolumn{1}{c}{-85.67}   & \multicolumn{1}{c}{0.0}        & \multicolumn{1}{c}{-29.91}   & 2.80$\times e^{-121}$         \\
5                     & \multicolumn{1}{c}{-83.69}   & \multicolumn{1}{c}{0.0}        & \multicolumn{1}{c}{-29.38}   & 1.72$\times e^{-118}$         \\
6                     & \multicolumn{1}{c}{-81.37}   & \multicolumn{1}{c}{0.0}        & \multicolumn{1}{c}{-29.38}   & 1.01$\times e^{-114}$         \\
7                     & \multicolumn{1}{c}{-78.87}   & \multicolumn{1}{c}{6.84$\times e^{-319}$}        & \multicolumn{1}{c}{-28.66}   & 3.11$\times e^{-110}$         \\
8                     & \multicolumn{1}{c}{-76.33}   & \multicolumn{1}{c}{3.89$\times e^{-311}$}        & \multicolumn{1}{c}{-27.81}   & 1.38$\times e^{-104}$         \\
9                     & \multicolumn{1}{c}{-73.58}   & \multicolumn{1}{c}{1.68$\times e^{-302}$}        & \multicolumn{1}{c}{-25.29}   & 6.51$\times e^{-97}$         \\
10                     & \multicolumn{1}{c}{-70.45}   & \multicolumn{1}{c}{2.25$\times e^{-292}$}        & \multicolumn{1}{c}{-23.34}   & 1.69$\times e^{-86}$         \\
\hline
\end{tabular}
\caption{T-test between LEAP and GRU-$\Delta t$/NCDE on MuJoCo}\label{tbl:stat_significance}
\end{table}

\paragraph{Statistical Significance} We test the statistical significance on MuJoCo using the paired t-test with a significance threshold of $0.01$. Its null hypothesis and alternative hypothesis are set as follows:\begin{align}
H_0 : \mathcal{E}_{ours} \geq \mathcal{E}_{base}, \quad H_1 : \mathcal{E}_{ours} < \mathcal{E}_{base},
\end{align} where $\mathcal{E}$ is error between predicted values and true values measured in the $L^2$ vector norm. We do paired t-tests with every baseline on Table~\ref{tbl:mujoco}, and p-values are under 0.01 in every case, confirming our alternative hypothesis with high statistical significance, and rejecting the null hypothesis. The results of the paired t-test against GRU-$\Delta t$ and NCDE are in Table~\ref{tbl:stat_significance}, showing extremely low p-values. We report for each forecasting horizon --- recall that we predict next 10 values in MuJoCo.

\subsection{Visualization}
\begin{figure}
    \centering
    \subfigure[]{\includegraphics[width=0.49\columnwidth]{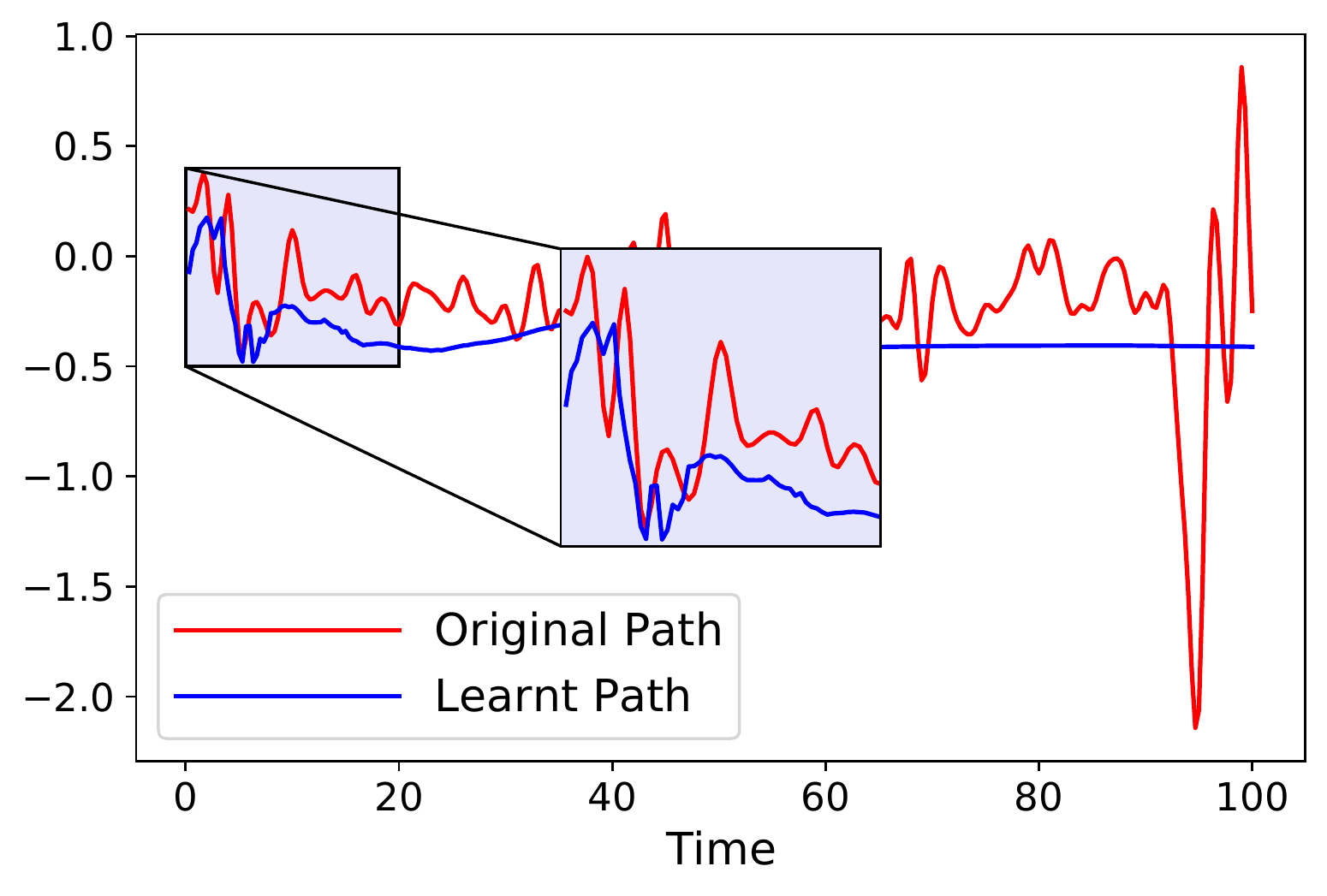}} \hfill
    \subfigure[]{\includegraphics[width=0.49\columnwidth]{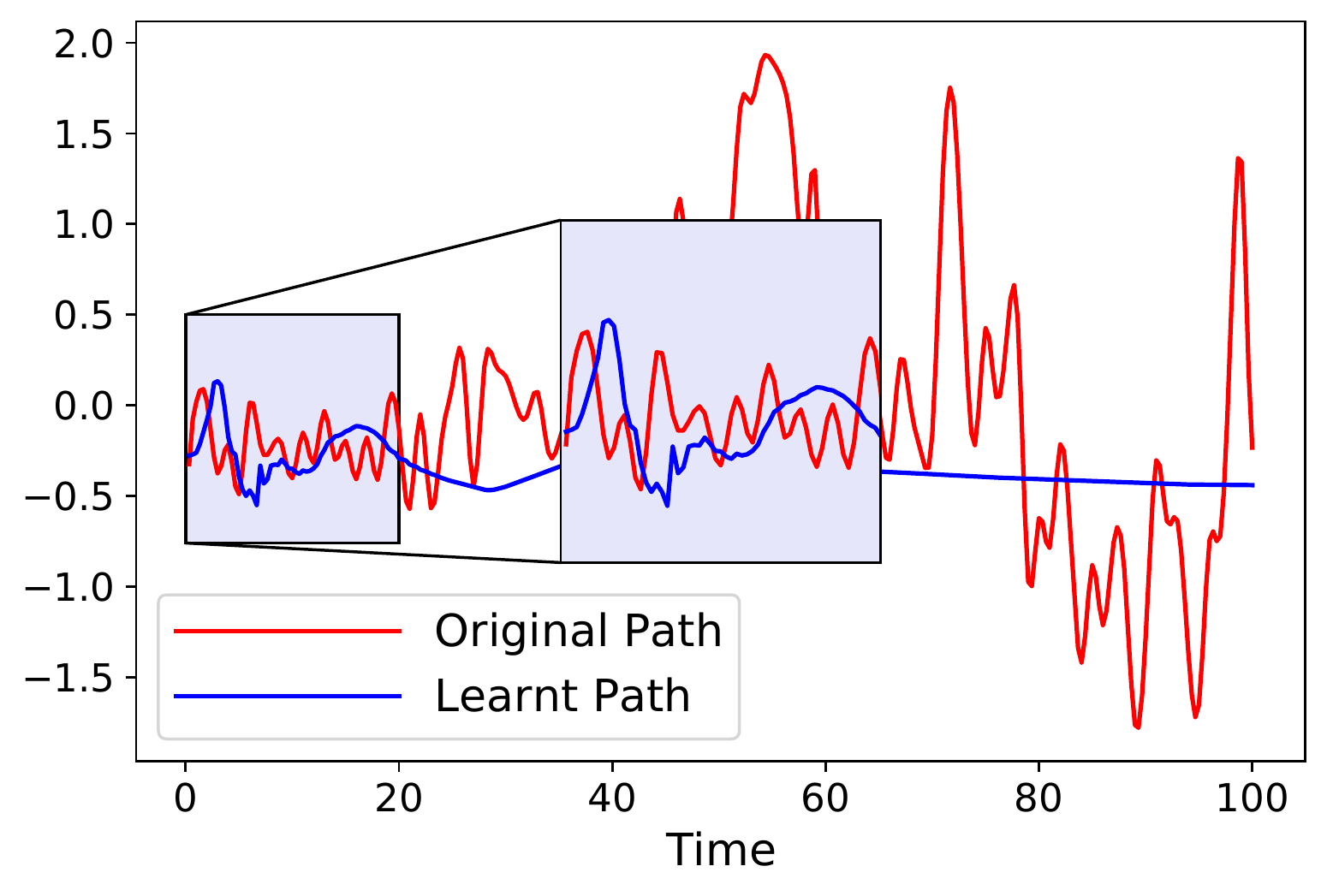}}\\
    \caption{Examples of the original path $X$ and learnt path $Y$ in Speech Commands.}
    \label{fig:speech_commands_path}
\end{figure}

\begin{figure}
    \centering
    \subfigure[]{\includegraphics[width=0.49\columnwidth]{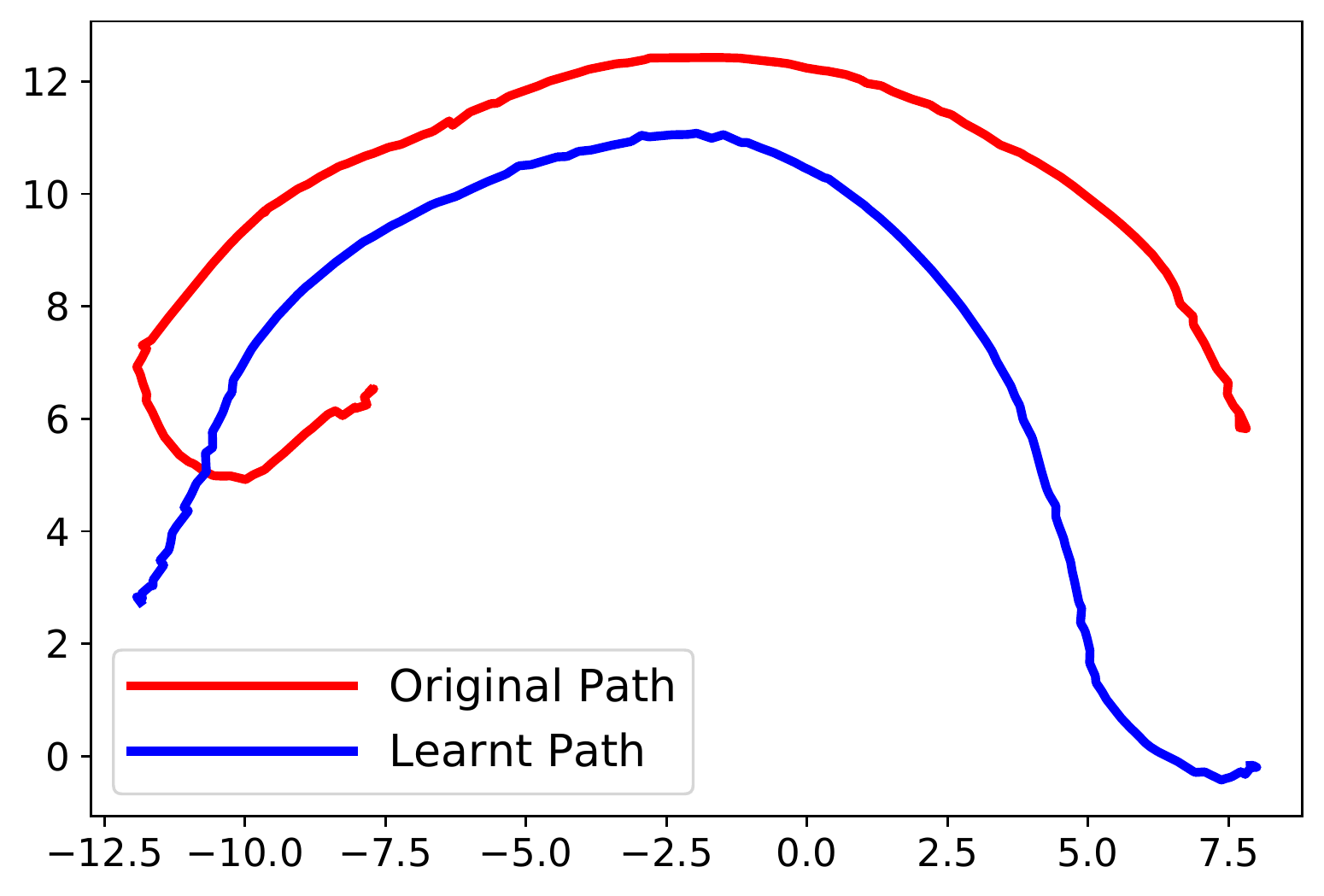}}\hfill
    \subfigure[]{\includegraphics[width=0.49\columnwidth]{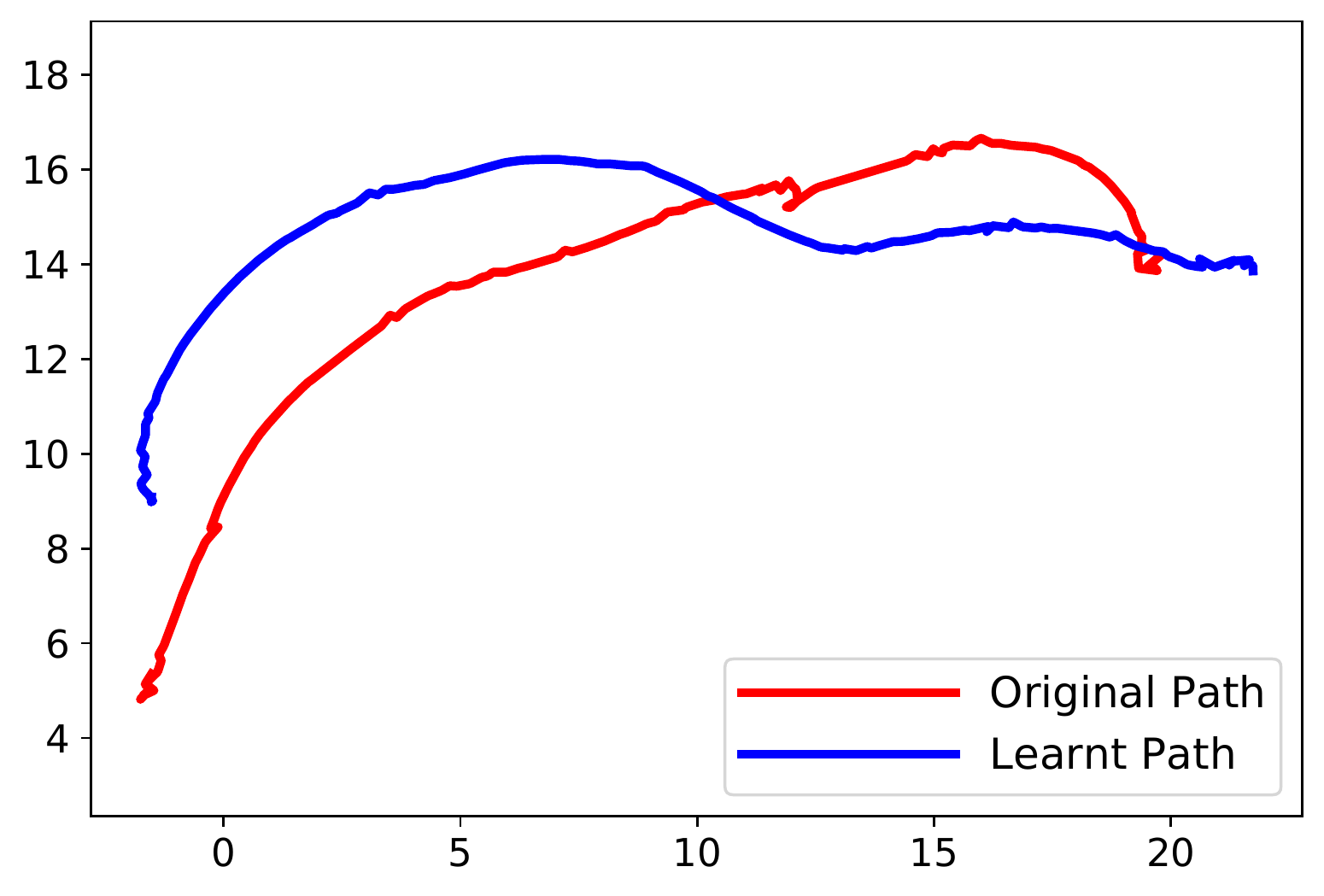}}
    \caption{Examples of the original path $X$ and learnt path $Y$ in MuJoCo using UMAP. More figures are in Appendix.}
    \label{fig:umap}
\end{figure}

We compare the original path $X$ and the learnt (or fine-tuned) path $Y$ in Fig.~\ref{fig:speech_commands_path}. According to Fig.~\ref{fig:speech_commands_path}, we select an element of a time-series sample in Speech Commands and visualize the original path and the learnt path on the selected element (dimension). Our MLE training to match them successfully ensures that they are well align when $t < 50$ for Fig.~\ref{fig:speech_commands_path} (a). However, the learnt path in blue can be considered a smoothed (fine-tuned) version of $X$ in the visualization. One remarkable point is that the learnt path becomes flat when $t \geq 50$. As shown in Eq.~\ref{eq:top}, the flat $Y$ curve makes $\frac{d Y(t)}{dt} \approx 0$ and $\mathbf{z}(t)$ is not much changed after $t=50$. In case of Fig.~\ref{fig:speech_commands_path} (b), same pattern is shown at $t < 60$. 

Fig.~\ref{fig:umap} shows other visualization in our datasets. For this, we use a different visualization method, UMAP~\cite{NBC2020}. This method is a lower-dimensional projection algorithm and each time-series sample is projected onto a 2-dim space. As shown in Fig.~\ref{fig:umap}, our learnt paths are quite similar to the original paths but with little variation. From those facts, we can see that LEAP has different degrees of learning paths depending on datasets for enhancing the task performance. 

\begin{table}[t]
\setlength{\tabcolsep}{1pt}
\scriptsize
\centering

\begin{tabular}{ccccc}
\hline
\multirow{3}{*}{Model}           & \multicolumn{4}{c}{Datasets} \\ \cline{2-5}
 & \multicolumn{2}{c}{Character Trajectories} & \multicolumn{2}{c}{MuJoCo} \\ \cline{2-5}
 & Test Accuracy           & Memory& Test MSE & Memory \\ \hline
RNN             & 0.311 ± 0.038           & 52.2  & 0.051 ± 0.001             & 409.1  \\
LSTM            & 0.791 ± 0.083           & 48.6  & 0.064 ± 0.001             & 411.2  \\
GRU             & 0.844 ± 0.089           & 54.8  & 0.053 ± 0.000             & 439.5 \\
GRU-$\Delta t$  & 0.834 ± 0.078           & 16.5  & 0.142 ± 0.020             & 532.9  \\
GRU-D           & 0.896 ± 0.030           & 17.8  & 0.471 ± 0.038             & 569.1  \\
GRU-ODE         & 0.878 ± 0.051           & 1.51  & 0.821 ± 0.027             & 146.2  \\
ODE-RNN         & 0.827 ± 0.048           & 15.5  & 0.234 ± 0.211             & 146.3  \\
Latent-ODE      & 0.942 ± 0.008           & 181   & 0.049 ± 0.007             & 314.9  \\
Augmented-ODE   & 0.970 ± 0.009           & 186   & 0.045 ± 0.004             & 286.1  \\
ACE-NODE        & 0.891 ± 0.001           & 113   & 0.046 ± 0.003             & 4,226  \\
NCDE            & 0.981 ± 0.002           & 1.38  & 0.028 ± 0.002             & 52.08  \\
ANCDE           & 0.986 ± 0.007           & 2.02  & 0.029 ± 0.003             & 79.22  \\
\hline
\textbf{LEAP} & \textbf{0.992 ± 0.001}  & 9.24 & \textbf{0.022 ± 0.002}    & 144.1 \\

\hline
\end{tabular}
\caption{Regular time-series prediction}\label{tbl:regular}
\end{table}

\paragraph{Regular Time-series}  As Character Trajectories and MuJoCo provide complete data without missing values, i.e., regular time-series, we conduct the tasks with the full information and their results are summarized in Table~\ref{tbl:regular}. In both datasets, our method shows the best performance.
One more point is that our method's regular and irregular forecasting results are the same for MuJoCo, which proves the strength of our method for irregular time-series. On top of that, accuracy of Character Trajectories with 50\% missing values is even higher than that of full informed data, while many other baselines --- ANCDE, ACE-NODE, Augmented-ODE, and so on --- show lower performances compared to score with 50\% missing data.

\section{Conclusions}


How to interpolate the input discrete time-series and create a continuous path is an important topic in NCDEs. In this work, we presented a method to learn how to interpolate from data (rather than relying on existing interpolation algorithms). To this end, we designed an encoder-decoder architecture and its special training method. We conducted a diverse set of experiments based on four datasets and twelve baselines, ranging from irregular/regular classification to forecasting. Our method, LEAP, clearly outperforms existing methods in almost all cases.

\section*{Acknowledgement}
Noseong Park is the corresponding author. 
This work was supported by the Yonsei University Research Fund of 2021, and the Institute of Information \& Communications Technology Planning \& Evaluation (IITP) grant funded by the Korean government (MSIT) (No. 2020-0-01361, Artificial Intelligence Graduate School Program (Yonsei University), and (No.2022-0-00857, Development of AI/data-based financial/economic digital twin platform,10\%) and (No.2022-0-00113, Developing a Sustainable Collaborative Multi-modal Lifelong Learning Framework, 45\%),(2022-0-01032, Development of Collective Collaboration Intelligence Framework for Internet of Autonomous Things, 45\%).



\bibliographystyle{aaai23}
\bibliography{aaai23}

\begin{thebibliography}{37}
\providecommand{\natexlab}[1]{#1}

\bibitem[{Bagnall et~al.(2018)Bagnall, Dau, Lines, Flynn, Large, Bostrom,
  Southam, and Keogh}]{bagnall2018uea}
Bagnall, A.; Dau, H.~A.; Lines, J.; Flynn, M.; Large, J.; Bostrom, A.; Southam,
  P.; and Keogh, E. 2018.
\newblock The UEA multivariate time series classification archive, 2018.
\newblock \emph{arXiv preprint arXiv:1811.00075}.

\bibitem[{Bai et~al.(2020)Bai, Yao, Li, Wang, and Wang}]{NEURIPS2020_ce1aad92}
Bai, L.; Yao, L.; Li, C.; Wang, X.; and Wang, C. 2020.
\newblock Adaptive Graph Convolutional Recurrent Network for Traffic
  Forecasting.
\newblock In Larochelle, H.; Ranzato, M.; Hadsell, R.; Balcan, M.~F.; and Lin,
  H., eds., \emph{NeurIPS}, volume~33, 17804--17815.

\bibitem[{Black and Scholes(1973)}]{10.2307/1831029}
Black, F.; and Scholes, M. 1973.
\newblock The Pricing of Options and Corporate Liabilities.
\newblock \emph{Journal of Political Economy}, 81(3): 637--654.

\bibitem[{Brouwer et~al.(2019)Brouwer, Simm, Arany, and
  Moreau}]{debrouwer2019gruodebayes}
Brouwer, E.~D.; Simm, J.; Arany, A.; and Moreau, Y. 2019.
\newblock GRU-ODE-Bayes: Continuous modeling of sporadically-observed time
  series.
\newblock In \emph{NeurIPS}.

\bibitem[{Che et~al.(2016)Che, Purushotham, Cho, Sontag, and
  Liu}]{che2016recurrent}
Che, Z.; Purushotham, S.; Cho, K.; Sontag, D.; and Liu, Y. 2016.
\newblock Recurrent Neural Networks for Multivariate Time Series with Missing
  Values.
\newblock arXiv:1606.01865.

\bibitem[{Chen et~al.(2018)Chen, Rubanova, Bettencourt, and
  Duvenaud}]{NIPS2018_7892}
Chen, R. T.~Q.; Rubanova, Y.; Bettencourt, J.; and Duvenaud, D.~K. 2018.
\newblock Neural Ordinary Differential Equations.
\newblock In \emph{NeurIPS}.

\bibitem[{Chen, Segovia-Dominguez, and Gel(2021)}]{chen2021ZGCNET}
Chen, Y.; Segovia-Dominguez, I.; and Gel, Y.~R. 2021.
\newblock Z-GCNETs: Time Zigzags at Graph Convolutional Networks for Time
  Series Forecasting.
\newblock \emph{arXiv preprint arXiv:2105.04100}.

\bibitem[{Choi et~al.(2022)Choi, Choi, Hwang, and Park}]{choi2022STGNCDE}
Choi, J.; Choi, H.; Hwang, J.; and Park, N. 2022.
\newblock Graph Neural Controlled Differential Equations for Traffic
  Forecasting.
\newblock In \emph{AAAI}.

\bibitem[{Chung et~al.(2014)Chung, Gulcehre, Cho, and
  Bengio}]{chung2014empirical}
Chung, J.; Gulcehre, C.; Cho, K.; and Bengio, Y. 2014.
\newblock Empirical evaluation of gated recurrent neural networks on sequence
  modeling.
\newblock \emph{arXiv preprint arXiv:1412.3555}.

\bibitem[{Dupont, Doucet, and Teh(2019)}]{NIPS2019_8577}
Dupont, E.; Doucet, A.; and Teh, Y.~W. 2019.
\newblock Augmented Neural ODEs.
\newblock In \emph{NeurIPS}.

\bibitem[{Fang et~al.(2021)Fang, Long, Song, and Xie}]{fang2021STODE}
Fang, Z.; Long, Q.; Song, G.; and Xie, K. 2021.
\newblock Spatial-Temporal Graph ODE Networks for Traffic Flow Forecasting.
\newblock \emph{arXiv preprint arXiv:2106.12931}.

\bibitem[{Fawaz et~al.(2019)Fawaz, Forestier, Weber, Idoumghar, and
  Muller}]{fawaz2019deep}
Fawaz, H.~I.; Forestier, G.; Weber, J.; Idoumghar, L.; and Muller, P.-A. 2019.
\newblock Deep learning for time series classification: a review.
\newblock \emph{Data mining and knowledge discovery}, 33(4): 917--963.

\bibitem[{Fu(2011)}]{fu2011review}
Fu, T.-c. 2011.
\newblock A review on time series data mining.
\newblock \emph{Engineering Applications of Artificial Intelligence}, 24(1):
  164--181.

\bibitem[{Grathwohl et~al.(2019)Grathwohl, Chen, Bettencourt, Sutskever, and
  Duvenaud}]{grathwohl2019ffjord}
Grathwohl, W.; Chen, R. T.~Q.; Bettencourt, J.; Sutskever, I.; and Duvenaud, D.
  2019.
\newblock FFJORD: Free-form Continuous Dynamics for Scalable Reversible
  Generative Models.
\newblock In \emph{ICLR}.

\bibitem[{Guo et~al.(2019)Guo, Lin, Feng, Song, and Wan}]{guo2019astgcn}
Guo, S.; Lin, Y.; Feng, N.; Song, C.; and Wan, H. 2019.
\newblock Attention Based Spatial-Temporal Graph Convolutional Networks for
  Traffic Flow Forecasting.
\newblock \emph{AAAI}, 33(01): 922--929.

\bibitem[{Hochreiter and Schmidhuber(1997)}]{sepp1997long}
Hochreiter, S.; and Schmidhuber, J. 1997.
\newblock Long Short-term Memory.
\newblock \emph{Neural computation}, 9: 1735--80.

\bibitem[{Huang et~al.(2020)Huang, Huang, Liu, Dai, and Kong}]{huang2020lsgcn}
Huang, R.; Huang, C.; Liu, Y.; Dai, G.; and Kong, W. 2020.
\newblock LSGCN: Long Short-Term Traffic Prediction with Graph Convolutional
  Networks.
\newblock In \emph{IJCAI}, 2355--2361.

\bibitem[{Hutchinson(1990)}]{doi:10.1080/03610919008812866}
Hutchinson, M. 1990.
\newblock A stochastic estimator of the trace of the influence matrix for
  laplacian smoothing splines.
\newblock \emph{Communications in Statistics - Simulation and Computation},
  19(2).

\bibitem[{Hwang et~al.(2021)Hwang, Choi, Choi, Lee, Lee, and
  Park}]{hwang2021climate}
Hwang, J.; Choi, J.; Choi, H.; Lee, K.; Lee, D.; and Park, N. 2021.
\newblock Climate Modeling with Neural Diffusion Equations.
\newblock In \emph{ICDM}, 230--239.

\bibitem[{Jhin et~al.(2021{\natexlab{a}})Jhin, Jo, Kong, Jeon, and
  Park}]{jhin2021acenode}
Jhin, S.~Y.; Jo, M.; Kong, T.; Jeon, J.; and Park, N. 2021{\natexlab{a}}.
\newblock ACE-NODE: Attentive Co-Evolving Neural Ordinary Differential
  Equations.
\newblock In \emph{KDD}.

\bibitem[{Jhin et~al.(2023)Jhin, Jo, Kook, and Park}]{jhinjo2023leap}
Jhin, S.~Y.; Jo, M.; Kook, S.; and Park, N. 2023.
\newblock Learnable Path in Neural Controlled Differential Equations.

\bibitem[{Jhin et~al.(2021{\natexlab{b}})Jhin, Shin, Hong, Jo, Park, and
  Park}]{jhin2021ancde}
Jhin, S.~Y.; Shin, H.; Hong, S.; Jo, M.; Park, S.; and Park, N.
  2021{\natexlab{b}}.
\newblock Attentive Neural Controlled Differential Equations for Time-series
  Classification and Forecasting.

\bibitem[{Jordan, Sokol, and Park(2019)}]{jordan2019gated}
Jordan, I.~D.; Sokol, P.~A.; and Park, I.~M. 2019.
\newblock Gated recurrent units viewed through the lens of continuous time
  dynamical systems.
\newblock arXiv:1906.01005.

\bibitem[{Kidger et~al.(2020)Kidger, Morrill, Foster, and
  Lyons}]{NEURIPS2020_4a5876b4}
Kidger, P.; Morrill, J.; Foster, J.; and Lyons, T. 2020.
\newblock Neural Controlled Differential Equations for Irregular Time Series.
\newblock In \emph{NeurIPS}.

\bibitem[{Lyons, Caruana, and Lévy(2004)}]{lyons2004differential}
Lyons, T.; Caruana, M.; and Lévy, T. 2004.
\newblock \emph{Differential Equations Driven by Rough Paths}.
\newblock Springer.
\newblock École D'Eté de Probabilités de Saint-Flour XXXIV - 2004.

\bibitem[{Morrill et~al.(2021)Morrill, Kidger, Yang, and
  Lyons}]{morrill2021neural}
Morrill, J.; Kidger, P.; Yang, L.; and Lyons, T. 2021.
\newblock Neural Controlled Differential Equations for Online Prediction Tasks.
\newblock \emph{arXiv preprint arXiv:2106.11028}.

\bibitem[{Reinsel(2003)}]{reinsel2003elements}
Reinsel, G.~C. 2003.
\newblock \emph{Elements of multivariate time series analysis}.
\newblock Springer Science \& Business Media.

\bibitem[{Reiter(2005)}]{article}
Reiter, P.~J. 2005.
\newblock Using CART to Generate Partially Synthetic, Public Use Microdata.
\newblock \emph{Journal of Official Statistics}, 21: 441.

\bibitem[{Reyna et~al.(2019)Reyna, Josef, Seyedi, Jeter, Shashikumar,
  Brandon~Westover, Sharma, Nemati, and Clifford}]{9005736}
Reyna, M.~A.; Josef, C.; Seyedi, S.; Jeter, R.; Shashikumar, S.~P.;
  Brandon~Westover, M.; Sharma, A.; Nemati, S.; and Clifford, G.~D. 2019.
\newblock Early Prediction of Sepsis from Clinical Data: the
  PhysioNet/Computing in Cardiology Challenge 2019.
\newblock In \emph{CinC}, Page 1--Page 4.

\bibitem[{Rubanova, Chen, and Duvenaud(2019)}]{NEURIPS2019_42a6845a}
Rubanova, Y.; Chen, R. T.~Q.; and Duvenaud, D.~K. 2019.
\newblock Latent Ordinary Differential Equations for Irregularly-Sampled Time
  Series.
\newblock In \emph{NeurIPS}.

\bibitem[{Sainburg, McInnes, and Gentner(2020)}]{NBC2020}
Sainburg, T.; McInnes, L.; and Gentner, T.~Q. 2020.
\newblock Parametric UMAP: learning embeddings with deep neural networks for
  representation and semi-supervised learning.
\newblock \emph{ArXiv e-prints}.

\bibitem[{Song et~al.(2020)Song, Lin, Guo, and Wan}]{song2020stsgcn}
Song, C.; Lin, Y.; Guo, S.; and Wan, H. 2020.
\newblock Spatial-Temporal Synchronous Graph Convolutional Networks: A New
  Framework for Spatial-Temporal Network Data Forecasting.
\newblock \emph{AAAI}, 34(01): 914--921.

\bibitem[{Tassa et~al.(2018)Tassa, Doron, Muldal, Erez, Li, de~Las~Casas,
  Budden, Abdolmaleki, Merel, Lefrancq, Lillicrap, and
  Riedmiller}]{DBLP:journals/corr/abs-1801-00690}
Tassa, Y.; Doron, Y.; Muldal, A.; Erez, T.; Li, Y.; de~Las~Casas, D.; Budden,
  D.; Abdolmaleki, A.; Merel, J.; Lefrancq, A.; Lillicrap, T.~P.; and
  Riedmiller, M.~A. 2018.
\newblock DeepMind Control Suite.
\newblock \emph{CoRR}, abs/1801.00690.

\bibitem[{Warden(2018)}]{DBLP:journals/corr/abs-1804-03209}
Warden, P. 2018.
\newblock Speech Commands: {A} Dataset for Limited-Vocabulary Speech
  Recognition.
\newblock \emph{CoRR}, abs/1804.03209.

\bibitem[{Wu et~al.(2019)Wu, Pan, Long, Jiang, and Zhang}]{wu2019graphwavenet}
Wu, Z.; Pan, S.; Long, G.; Jiang, J.; and Zhang, C. 2019.
\newblock Graph WaveNet for Deep Spatial-Temporal Graph Modeling.
\newblock In \emph{IJCAI}, 1907--1913.

\bibitem[{Yu, Yin, and Zhu(2018)}]{bing2018stgcn}
Yu, B.; Yin, H.; and Zhu, Z. 2018.
\newblock Spatio-Temporal Graph Convolutional Networks: A Deep Learning
  Framework for Traffic Forecasting.
\newblock In \emph{IJCAI}, 3634--3640.

\bibitem[{Zhang et~al.(2019)Zhang, Song, Chen, Feng, Lumezanu, Cheng, Ni, Zong,
  Chen, and Chawla}]{zhang2019deep}
Zhang, C.; Song, D.; Chen, Y.; Feng, X.; Lumezanu, C.; Cheng, W.; Ni, J.; Zong,
  B.; Chen, H.; and Chawla, N.~V. 2019.
\newblock A deep neural network for unsupervised anomaly detection and
  diagnosis in multivariate time series data.
\newblock In \emph{AAAI}, volume~33, 1409--1416.

\end{thebibliography}

\clearpage

\appendix
\begin{appendices}
\begin{table}[t]
\scriptsize
\setlength{\tabcolsep}{1pt}
\centering
\caption{The best architecture of the CDE functions $k$ and $g$  for time series classification. \texttt{FC}, $\rho$, and $\xi$ stands for the fully-connected layer, the rectified linear unit (ReLU), and the hyperbolic tangent (tanh), respectively.}\label{tbl:f1}
\begin{tabular}{cccccccc}
\hline
{\color[HTML]{333333} }   &     & \multicolumn{2}{c}{{\color[HTML]{333333} Character Trajectories}} & \multicolumn{2}{c}{{\color[HTML]{333333} Speech Commands}}         & \multicolumn{2}{c}{{\color[HTML]{333333} PhysioNet Sepsis}}                 \\ \cline{3-8} 
\multirow{-2}{*}{{\color[HTML]{333333} Design}} & \multirow{-2}{*}{{\color[HTML]{333333} Layer}} & Input & Output & Input & Output  & Input & Output \\ \hline
\texttt{FC}& 1 & $32 \times $40  & $32 \times $100 & $1024 \times $90    & $1024 \times $40   & $1024 \times $49   & $1024 \times $39     \\

$\rho$(\texttt{FC}) & 2 & $32 \times $100  & $32 \times $100 & $1024 \times $40    & $1024 \times $40   & $1024 \times $39   & $1024 \times $39     \\
$\rho$(\texttt{FC}) & 3 & $32 \times $100  & $32 \times $100 & $1024 \times $40    & $1024 \times $40   & $1024 \times $39   & $1024 \times $39     \\

$\rho$(\texttt{FC}) & 4 & -  & - & $1024 \times $40    & $1024 \times $40   & -   & -     \\

$\xi$(\texttt{FC}) &  5  &$32 \times $100  & $32 \times $160 & $1024 \times $40   &$1024 \times $1890  &$1024 \times $39   & $1024 \times $3381    \\

\hline
\end{tabular}
\end{table}

\begin{table}[t]
\scriptsize
\setlength{\tabcolsep}{1pt}
\centering
\caption{The best architecture of the ODE function $f$ for time series classification}\label{tbl:f2}
\begin{tabular}{cccccccc}
\hline
{\color[HTML]{333333} }   &     & \multicolumn{2}{c}{{\color[HTML]{333333} Character Trajectories}} & \multicolumn{2}{c}{{\color[HTML]{333333} Speech Commands}}         & \multicolumn{2}{c}{{\color[HTML]{333333} PhysioNet Sepsis}}                 \\ \cline{3-8} 
\multirow{-2}{*}{{\color[HTML]{333333} Design}} & \multirow{-2}{*}{{\color[HTML]{333333} Layer}} & Input & Output & Input & Output  & Input & Output \\ \hline

$\rho$(\texttt{FC}) & 1 & $32 \times $40  & $32 \times $128 & $1024 \times $90    & $1024 \times $128   & $1024 \times $49   & $1024 \times $128     \\

$\xi$(\texttt{FC}) &  2  &$32 \times $128  & $32 \times $40 & $1024 \times $128   &$1024 \times $90  &$1024 \times $128   & $1024 \times $49    \\

\hline
\end{tabular}
\end{table}

\begin{table}[t]
\scriptsize
\setlength{\tabcolsep}{4pt}
\centering
\caption{The best architecture of the CDE function $k$ for time series forecasting}\label{tbl:f4}
\begin{tabular}{cccc}
\hline
Design & Layer & Input & Output  \\ 
\hline
\texttt{FC}&1 &  $1024 \times $80  & $1024 \times $50  \\ 
$\rho$(\texttt{FC}) &2 &  $1024 \times $50  & $1024 \times $50 \\ 
$\rho$(\texttt{FC}) &3 &  $1024 \times $50  & $1024 \times $50 \\ 
$\rho$(\texttt{FC}) &4 &  $1024 \times $50  & $1024 \times $50 \\ 
$\rho$(\texttt{FC}) &5 &  $1024 \times $50  & $1024 \times $50 \\ 
$\xi$(\texttt{FC}) &6 &  $1024 \times $50  & $1024 \times $1120 \\

\hline
\end{tabular}

\hfill
\scriptsize
\setlength{\tabcolsep}{4pt}
\centering
\caption{The best architecture of the CDE function $g$ for time series forecasting , $\varepsilon$ stands for the exponential linear unit (ELU)} \label{tbl:f4}
\begin{tabular}{cccc}
\hline
Design & Layer & Input & Output  \\ 
\hline
 \texttt{FC}&1 & $1024 \times $80  & $1024 \times $50  \\ 
$\varepsilon$(\texttt{FC}) &2 &  $1024 \times $50  & $1024 \times $50 \\ 
$\varepsilon$(\texttt{FC}) &3 &  $1024 \times $50  & $1024 \times $50 \\ 
$\varepsilon$(\texttt{FC}) &4 &  $1024 \times $50  & $1024 \times $50 \\
$\varepsilon$(\texttt{FC}) &5 &  $1024 \times $50  & $1024 \times $50 \\
$\varepsilon$(\texttt{FC}) &6 &  $1024 \times $50  & $1024 \times $50 \\
$\xi$(\texttt{FC}) &7 &  $1024 \times $50 & $1024 \times $1120\\
 
\hline
\end{tabular}
\end{table}

\begin{table}[t]
\scriptsize
\setlength{\tabcolsep}{4pt}
\centering
\caption{The best architecture of the ODE function $f$ for time series forecasting}\label{tbl:f5}
\begin{tabular}{cccc}
\hline
Design & Layer & Input & Output  \\ 
\hline

$\rho$(\texttt{FC}) &1 &  $1024 \times $80  & $1024 \times $40 \\ 
$\xi$(\texttt{FC}) &2 &  $1024 \times $40 & $1024 \times $80\\
 
\hline
\end{tabular}
\end{table}

\section{Hutchinson's Trace Estimator}\label{sec:hutchinson}
The exact log-density under ODE-based models can be computed through the instantaneous change of variable formula ~\cite{NEURIPS2018_69386f6b} as follows:
\begin{align}\label{eq:logDensity}
   \log p(\mathbf{z}(t_1)) = \log p(\mathbf{z}(t_0)) - \int^{t_1}_{t_0} \text{Tr}(\frac{\partial f}{\partial \mathbf{z}(t)}) dt
\end{align} where $\mathbf{z}(t_0)\sim p_{z(t_0)}(\mathbf{z}(t_0))$ is a base distribution and $f(\mathbf{z}(t),t,\theta)$ is a parametric function of the ODE. On top of that, the entire ODE formulation should be an homeomorphic function.

According to~\cite{grathwohl2019ffjord}, in order to approximate its Jacobian matrix, $\text{Tr}({\partial f \over  \partial \mathbf{z}(t) })$ from Eq.~\eqref{eq:logDensity} requires polynomial-time computation, which is $\mathcal{O}\big(D^2\big)$. Even with the approximation trick, it costs $D$ evaluations of $f$, as each element of the Jacobian's diagonal necessitates computing a distinct derivative of $f$. To overcome this computational issue, one can use Hutchinson's trace estimator~\cite{doi:10.1080/03610919008812866}, which takes a double product of the matrix with a noise vector to compute an unbiased estimate of matrix's trace:
\begin{align}\label{eq:hutchinson}
    \text{Tr}(A) = E_{p(\mathbf{\epsilon})}[\mathbf{\epsilon}^T A \mathbf{\epsilon}]
\end{align} 
For any $D$-by-$D$ matrix $A$ and distribution $p(\mathbf{\epsilon})$ over $D$-dimensional vectors, where $Cov(\mathbf{\epsilon})= \mathbf{I}$ and $\mathbb{E}[\mathbf{\epsilon}] = \mathbf{0}$, Eq.~\eqref{eq:hutchinson} holds.

\section{Hyperparameters}

For the best outcome of baselines, we conduct hyperparameter search for them based on the recommended hyperparameter set from each papers. Considered hyperparameter sets for each datasets are as follows:
\begin{enumerate}
\item In Character Trajectories, we train for 200 epochs with a batch size of 32, and stop early if the train loss doesn't decrease for 50 epochs. We use a learning rate of ${\{1.0\times e^{-4}}, {5.0\times e^{-4}}, {1.0\times e^{-3}}, {5.0\times e^{-3}}\}$ and a hidden vector dimension of $\{10, 20, 40, 50, 60\}$. For RNN, LSTM, and GRU, we use a hidden vector dimension of 40.
\item In Speech Commands, we train for 200 epochs with a batch size of 1,024, and stop early if the train loss doesn't decrease for 50 epochs. We use a learning rate of $\{{1.0\times e^{-6}}, {5.0\times e^{-6}}, {1.0\times e^{-5}}, {5.0\times e^{-5}}\}$ and a hidden vector dimension of $\{60,90,120\}$. For RNN, LSTM, and GRU, we use a hidden vector dimension of 160.
\item In PhysioNet Sepsis, we train for 300 epochs with a batch size of 1,024, and stop early if the train loss doesn't decrease for 50 epochs. We use a learning rate of $\{{5.0\times e^{-5}}, {1.0\times e^{-4}}, {5.0\times e^{-4}}, {1.0\times e^{-3}}\}$ and a hidden vector dimension of $\{50,60,70\}$.
\item In MuJoCo, we train for 1,000 epochs with a batch size of 1,024, and stop early if the train loss doesn't decrease for 100 epochs. We use a learning rate of $\{{5.0\times e^{-5}}, {1.0\times e^{-4}}, {5.0\times e^{-4}}, {1.0\times e^{-3}}\}$ and a hidden vector dimension of $\{50,60,70\}$  For RNN, LSTM, and GRU, we use a hidden vector dimension of 180.
\end{enumerate}

\section{Best Hyperparameter for Our Model} \label{sec:best_param}

For the reproducibility of our research, we summarize the best hyperparameter and architecture configuration in Tables~\ref{tbl:f1} to~\ref{tbl:f5} for each dataset.
In the case of our model, we consider the following hyperparameter configurations: the number of layers in the ODE/CDE functions is $\{1,2,3,4,5,6\}$, the dimensionality of hidden vector $\mathbf{h}(t)$ is $\{20,30,40,50,80,100\}$, we use a learning rate of 
$\{{1.0\times e^{-4}}, {5.0\times e^{-4}}, {1.0\times e^{-3}}, {5.0\times e^{-3}}\}$. In Eq.~\eqref{eq:loss}, $\alpha$ and $\beta$ are considered among $\{{1.0\times e^{-6}}, {1.0\times e^{-5}}, {1.0\times e^{-4}}\}$. The other best hyperparameter set for each dataset is as follows:

\begin{enumerate}

\item In Character Trajectories, we train for 200 epochs with a batch size of 32, and stop early if the train loss doesn't decrease for 50 epochs. We use $\alpha, \beta$ of ${1.0 \times e^{-6}}$.
\item In Speech Commands, we train for 200 epochs with a batch size of 1,024, and stop early if the train loss doesn't decrease for 50 epochs. We use $\alpha, \beta$ of ${1.0 \times e^{-6}}$.
\item In PhysioNet Sepsis, we train for 300 epochs with a batch size of 1,024, and stop early if the train loss doesn't decrease for 50 epochs. Only for this dataset, we search a hidden vector dimension of $\{39, 49, 59\}$. We use $\alpha, \beta$ of ${1.0 \times e^{-6}}$. 
\item In MuJoCo, we train for 1,000 epochs with a batch size of 1,024, and stop early if the train loss doesn't decrease for 100 epochs. We use $\alpha, \beta$ of ${1.0 \times e^{-4}}$.

\end{enumerate}

\section{Ablation on Other Simplified Encoding Layers}
We conduct several modifications on GRU-ODE, ODE-RNN, NCDE, and LEAP. To clarify LEAP's performance, we add a few feed-forward layers to the plain baselines. To confirm the effectiveness, we remove the log-likelihood term from Eq.~\eqref{eq:loss}. Those results are not that much effective as reported in Tables~\ref{tbl:learnablelayer_1} and~\ref{tbl:learnablelayer_2}. Our proposed encoder-decoder architecture is better than the modified baseline designs at capturing key information from time series while interpolating it.

\begin{table}[t]
\setlength{\tabcolsep}{1pt}
\scriptsize
\centering
\caption{Ablation on Learnable layers in Classification task}\label{tbl:learnablelayer_1}
\begin{tabular}{ccccc} 
\hline
\multirow{4}{*}{Model} & \multicolumn{4}{c}{Dataset} \\
\cline{2-5}

& \multicolumn{2}{c}{Test Accuracy}& & \multicolumn{1}{c}{Test AUROC}  \\ 

 \cline{2-3} \cline{5-5}
 
& Character & Speech & & PhysioNet \\  
& Trajectories & Commands & & Sepsis\\ \hline

Learnable layers + GRU-ODE      &  0.875    &0.554  &&  0.863 \\
Learnable layers + ODE-RNN      &  0.942    &0.428  &&  0.894 \\
Learnable layers + NCDE         &  0.985    &0.921  &&  0.881  \\
LEAP without Likelihood         &  0.973    &0.921  &&  0.897 \\\hline
\textbf{LEAP}                    &  \textbf{0.992} &\textbf{0.922}&&\textbf{0.908}  \\ \hline

\end{tabular}
\end{table}
\begin{table}[t]
\setlength{\tabcolsep}{1pt}
\scriptsize
\centering
\caption{Ablation on Learnable layers in Forecasting task}\label{tbl:learnablelayer_2}
\begin{tabular}{ccc} 
\hline
\multirow{1}{*}{Model} & Test MSE  \\  \hline
Learnable layers + GRU-ODE      &  0.765    \\
Learnable layers + ODE-RNN      &  0.752    \\
Learnable layers + NCDE         &  0.037    \\
LEAP without Likelihood         &  0.024     \\\hline
\textbf{LEAP}                    &  \textbf{0.022}  \\ \hline

\end{tabular}
\end{table}

\begin{table}[t]
\scriptsize
\centering
\caption{Model size ablation study in PhysioNet Sepsis} \label{tbl:ablsepsis}
\begin{tabular}{cccccc} \hline
Model & Test AUROC & Memory Usage (MB) \\ \hline
NCDE (small)                & 0.880              & 244      \\
\textbf{LEAP (small)}       & 0.900              & 275      \\ 
NCDE (medium)               & 0.879              & 304      \\
\textbf{LEAP (medium)}      & \textbf{0.908}     & 306      \\
NCDE (large)                & 0.874              & 406      \\
\textbf{LEAP (large)}       & 0.894              & 406      \\ \hline
\end{tabular}
\end{table}

\section{Ablation on the Model Size}
Table~\ref{tbl:ablsepsis} shows the AUROC score of various models in PhysioNet Sepsis. As shown, too small and too large models in terms of the number of parameters show sub-optimal scores for our LEAP. This pattern can be observed in other datasets as well. Moreover, whatever size LEAP is in, LEAP always shows a better AUROC score than NCDE, which verifies the high effectiveness and robustness of LEAP.

\section{Visualization}\label{sec:appendix_vis}
\begin{figure*}[ht]
    \centering
    \subfigure[Character Trajectories]{\includegraphics[width=0.48\columnwidth]{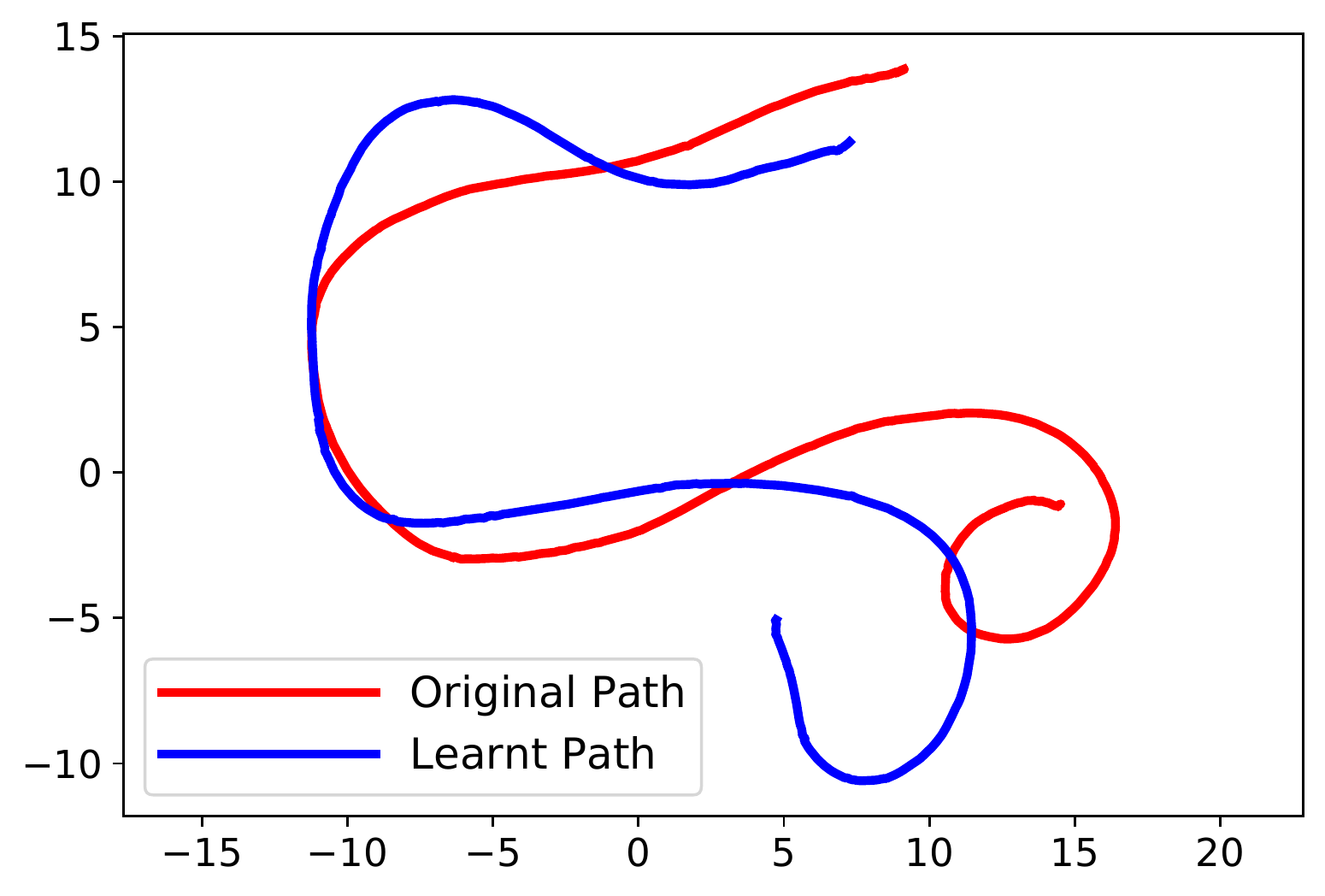}}
    \subfigure[Character Trajectories]{\includegraphics[width=0.48\columnwidth]{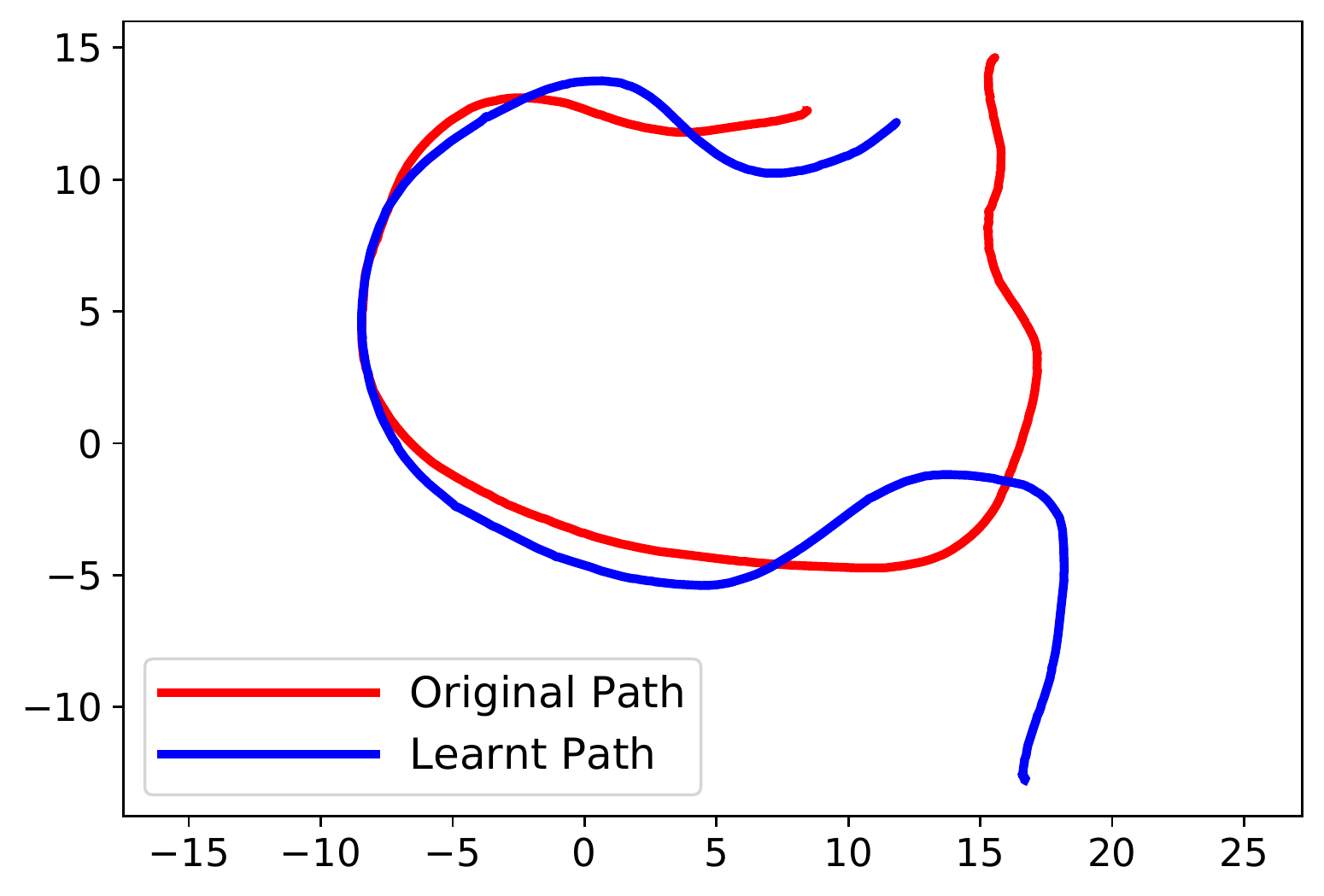}}
    \subfigure[Character Trajectories]{\includegraphics[width=0.48\columnwidth]{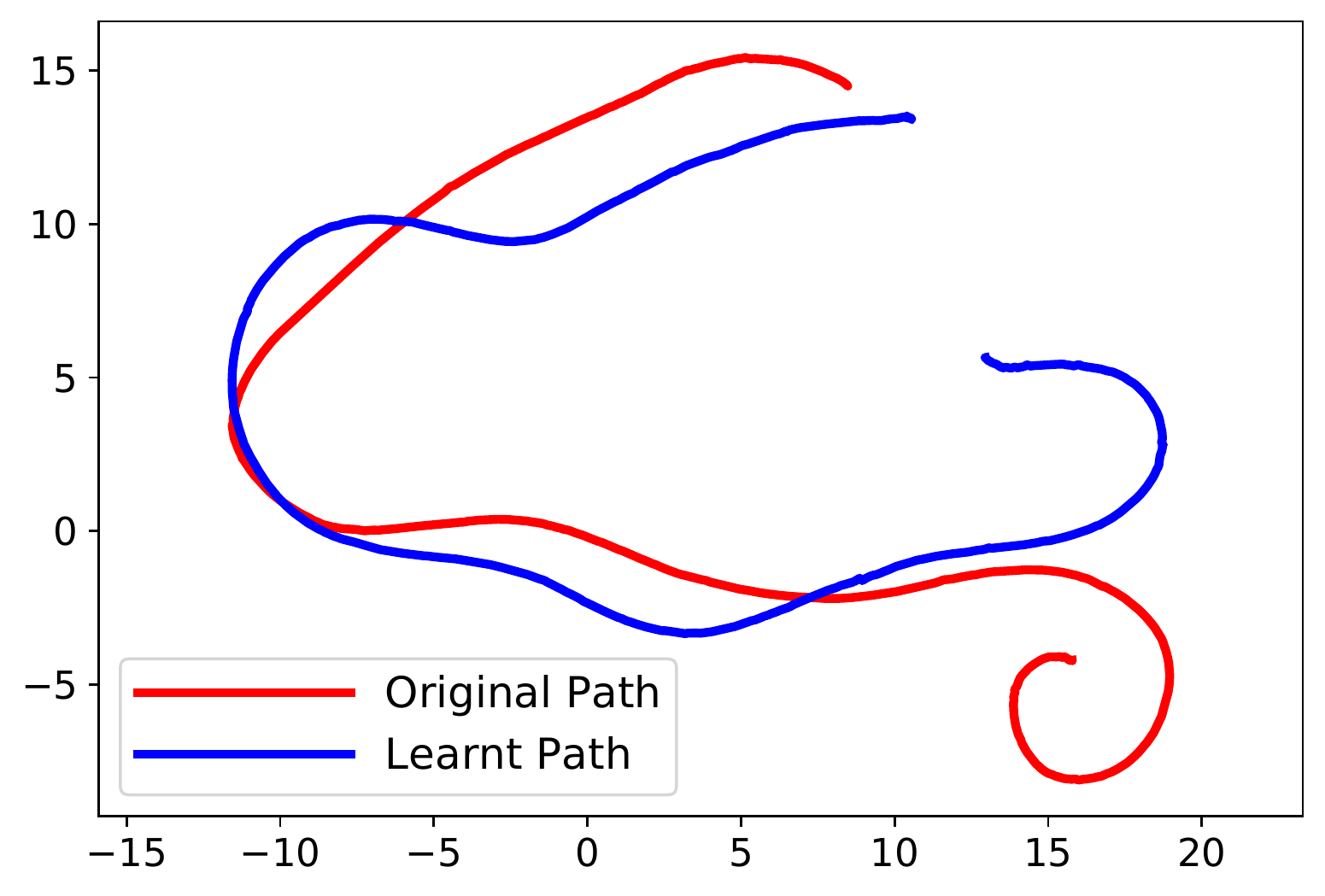}}
    \subfigure[Character Trajectories]{\includegraphics[width=0.48\columnwidth]{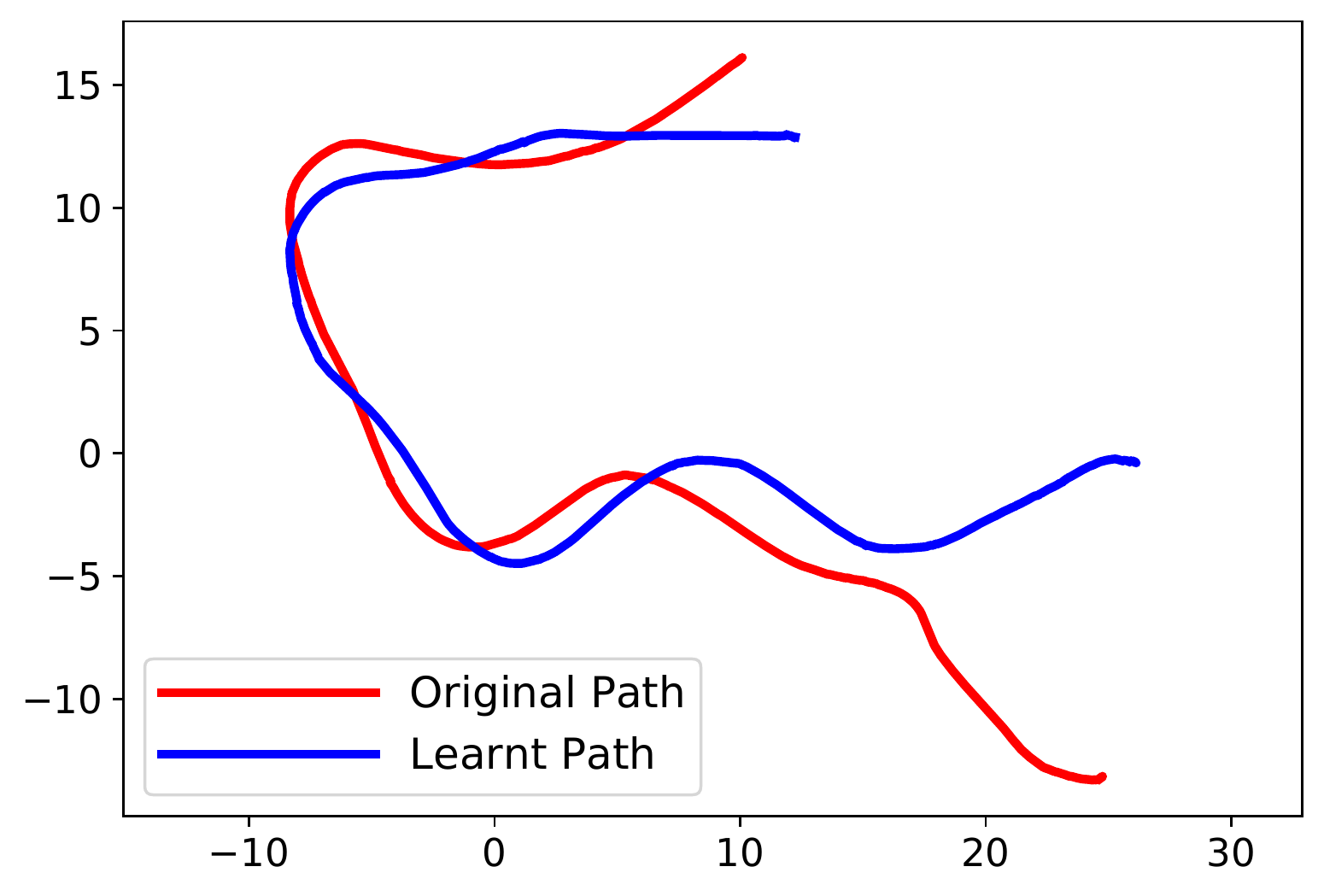}}
    
    \subfigure[PhysioNet Sepsis]{\includegraphics[width=0.48\columnwidth]{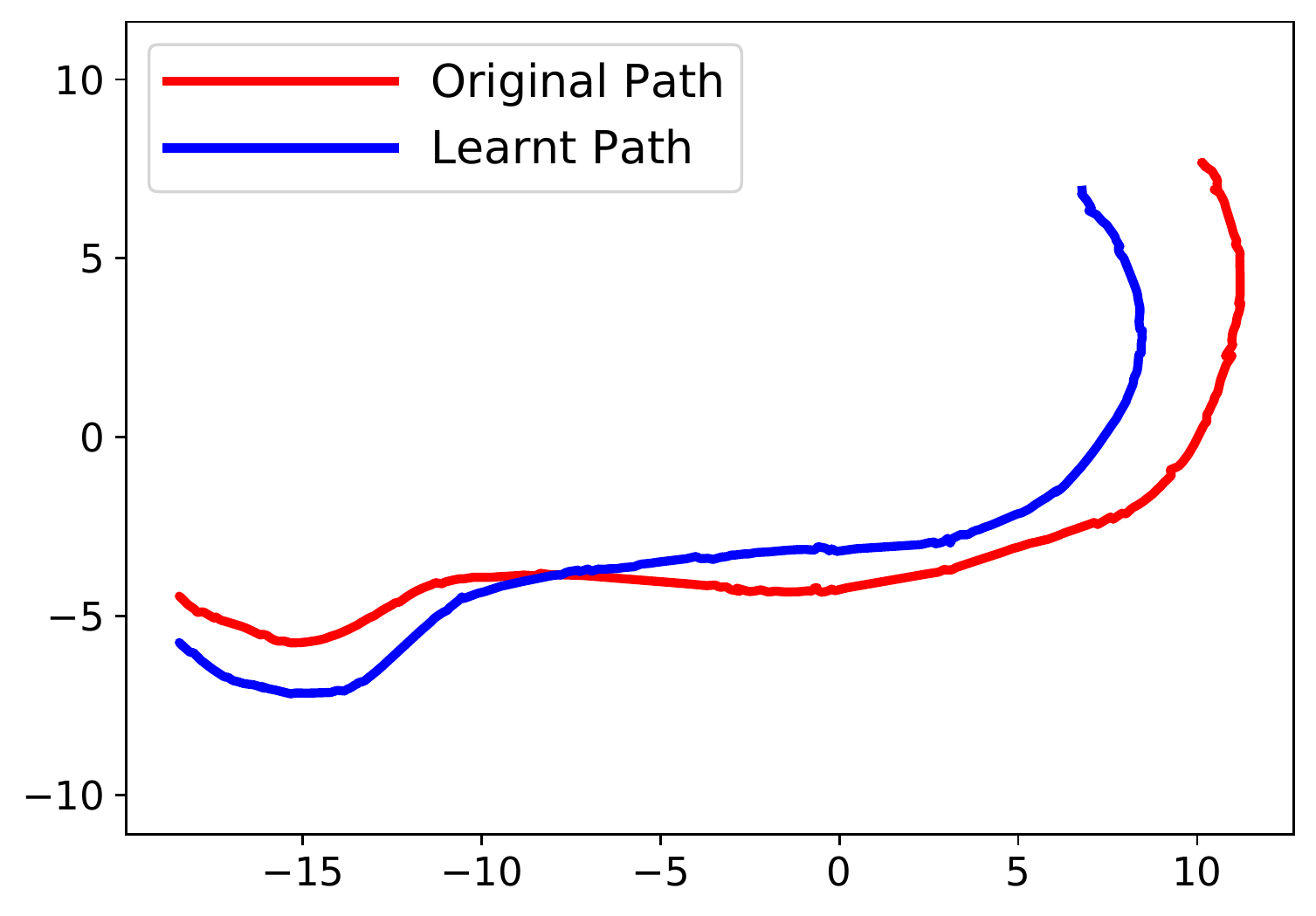}} 
    \subfigure[PhysioNet Sepsis]{\includegraphics[width=0.48\columnwidth]{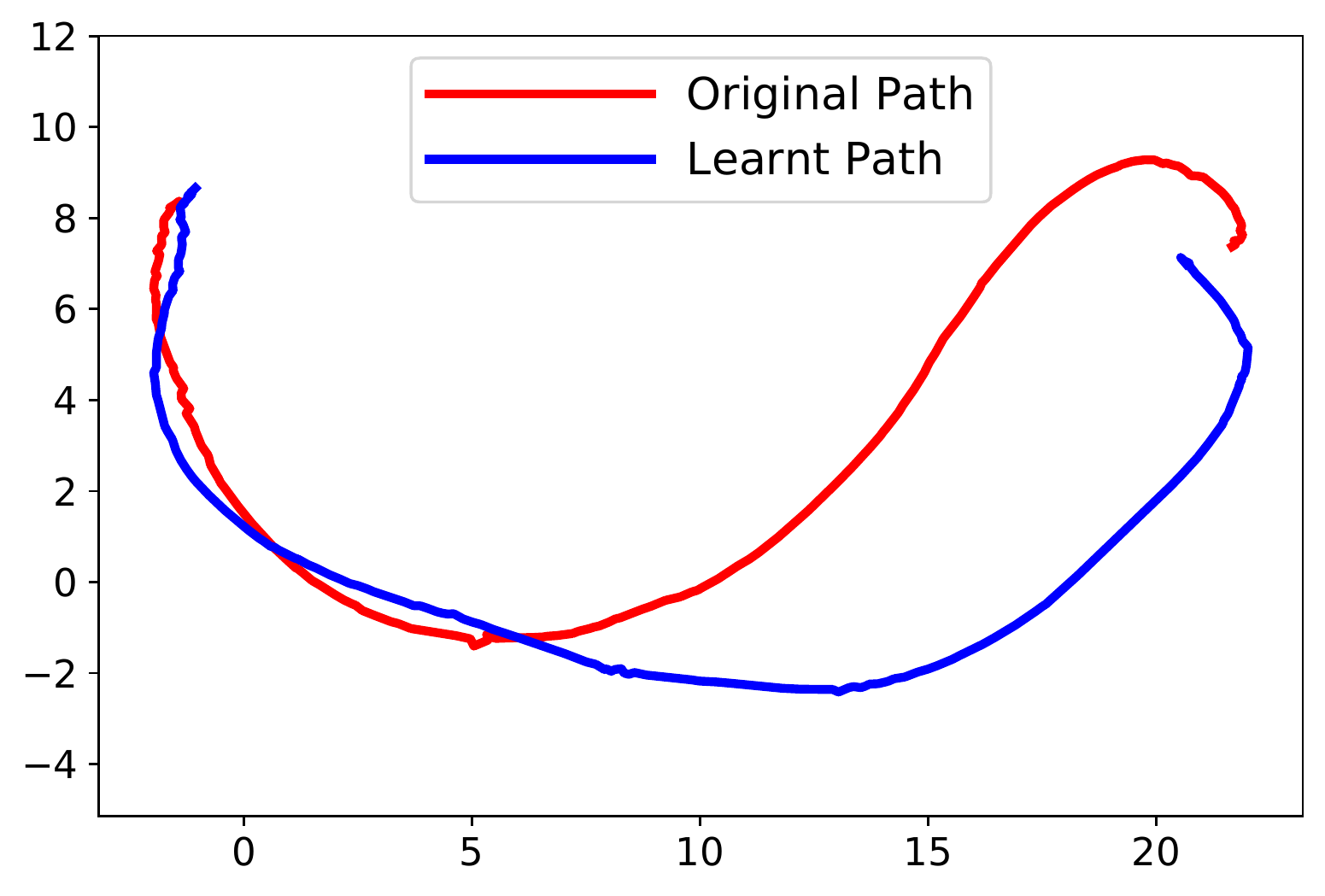}}
    \subfigure[PhysioNet Sepsis]{\includegraphics[width=0.48\columnwidth]{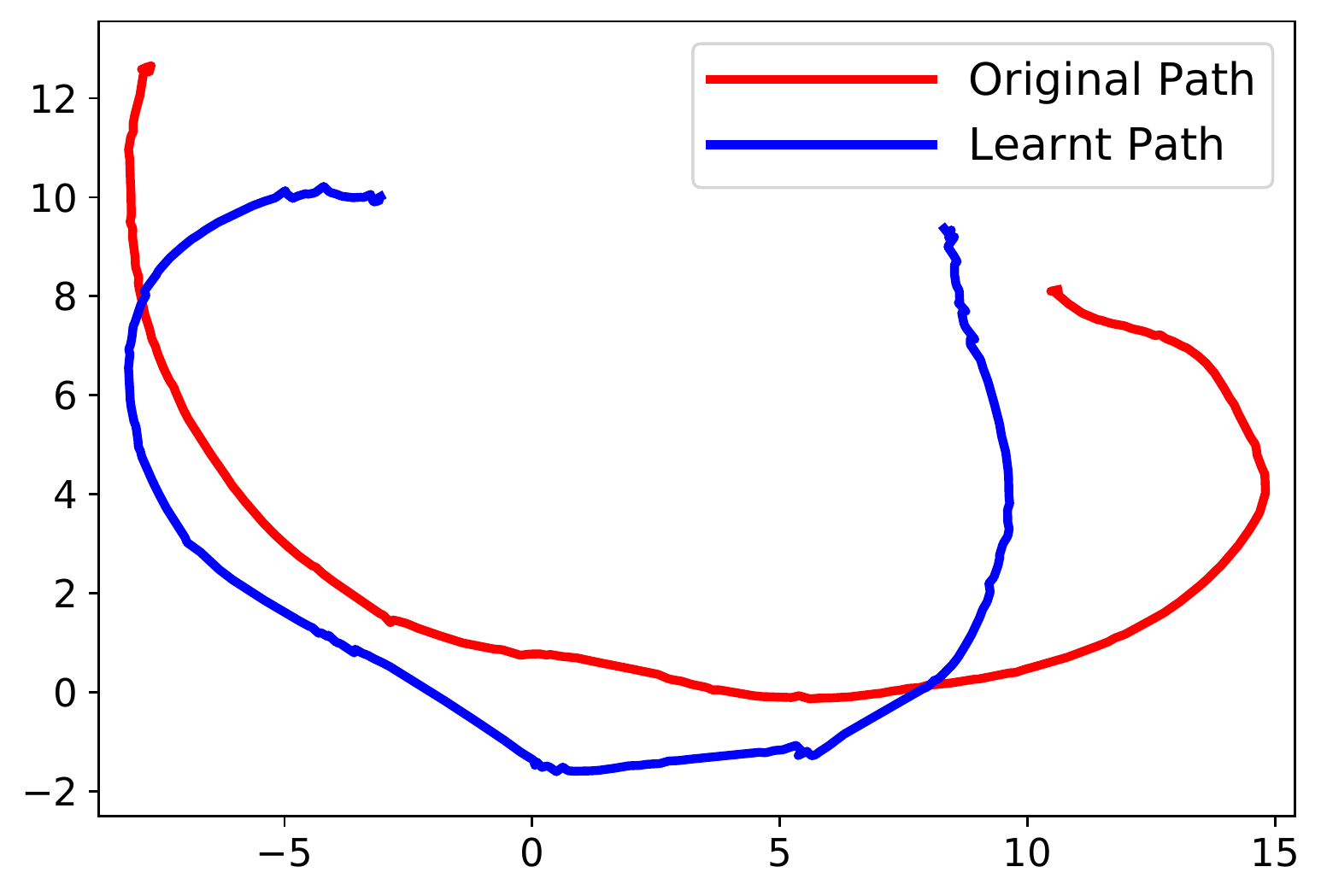}}
    \subfigure[PhysioNet Sepsis]{\includegraphics[width=0.48\columnwidth]{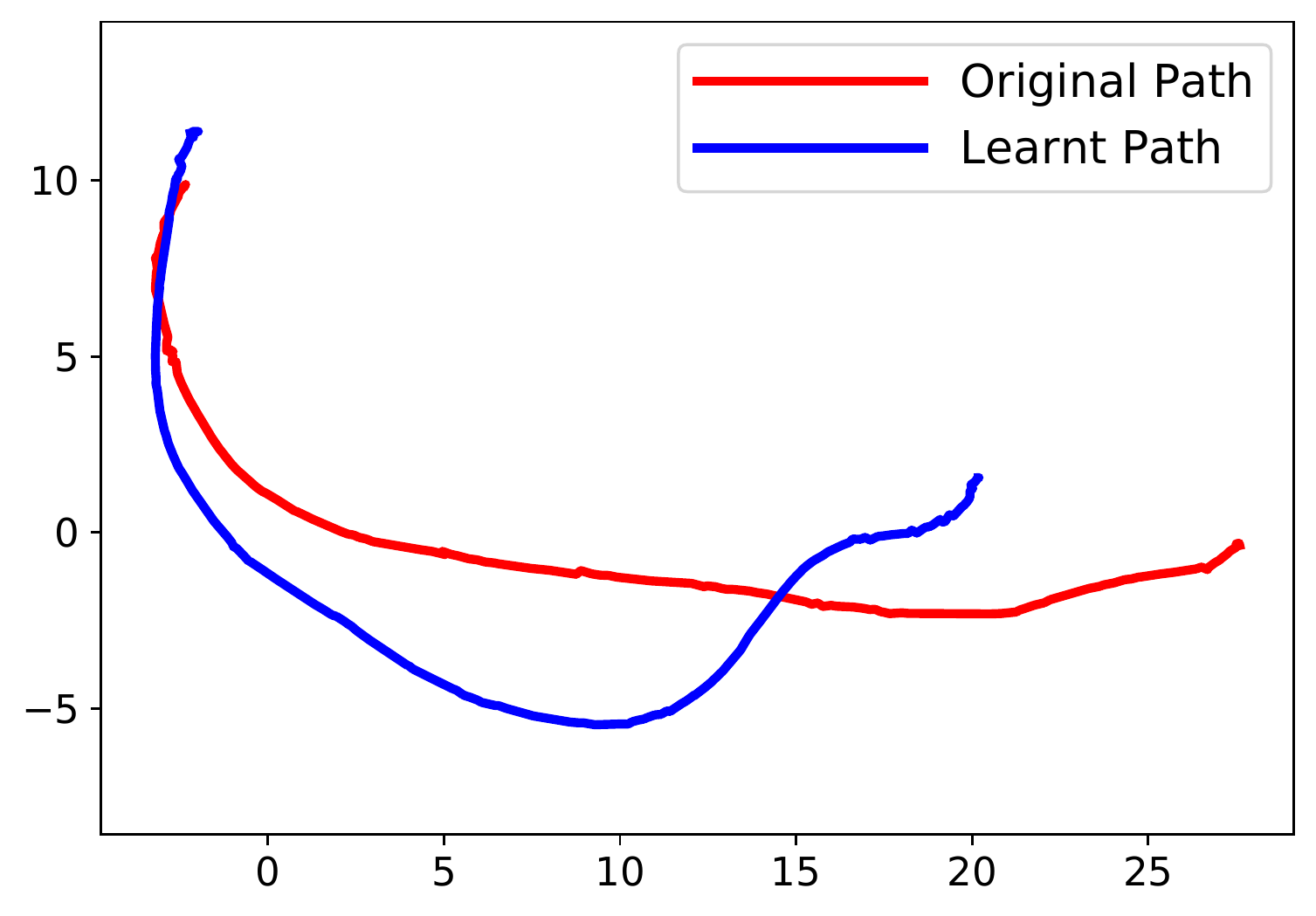}}
    \subfigure[Speech Commands]{\includegraphics[width=0.48\columnwidth]{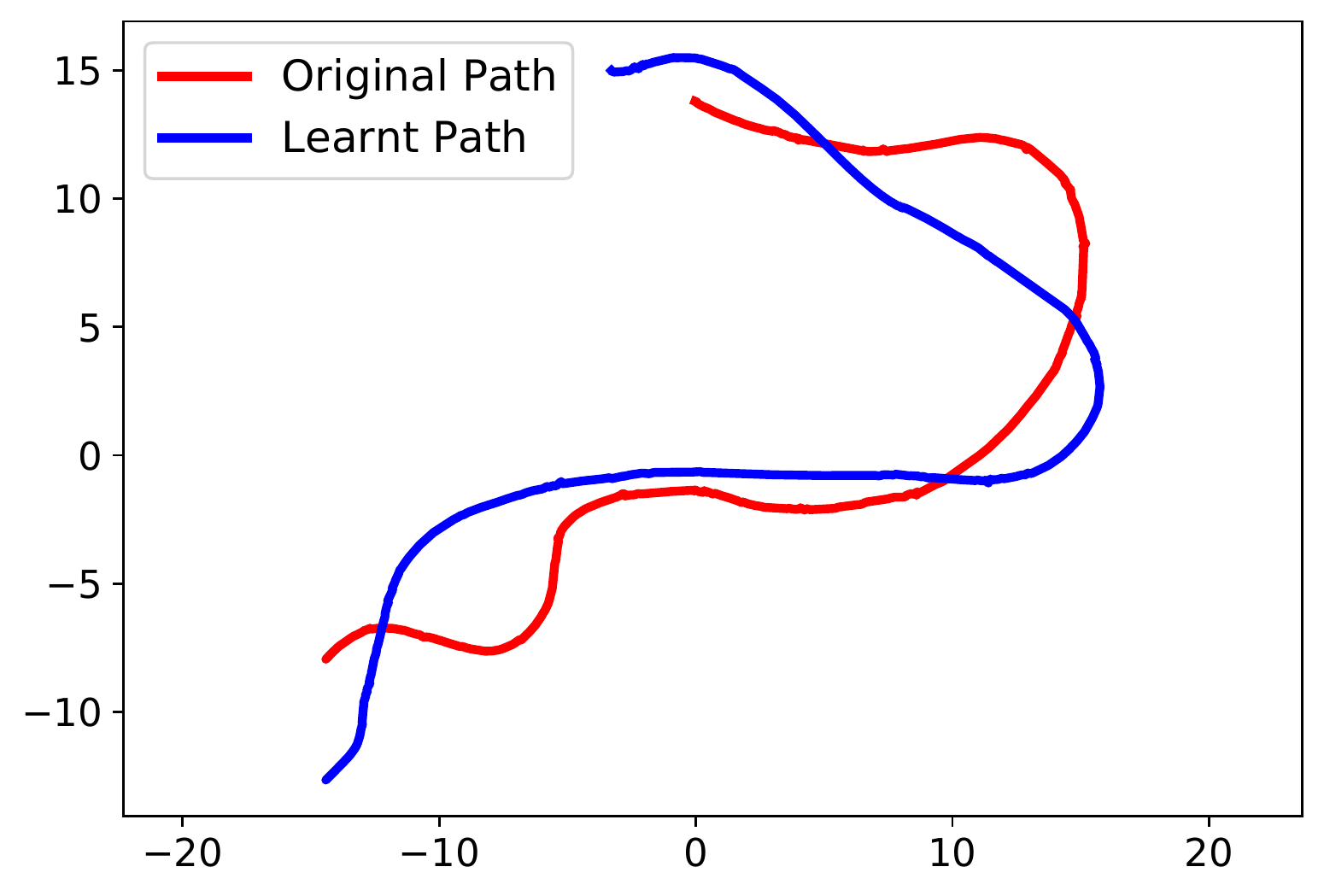}}
    \subfigure[Speech Commands]{\includegraphics[width=0.48\columnwidth]{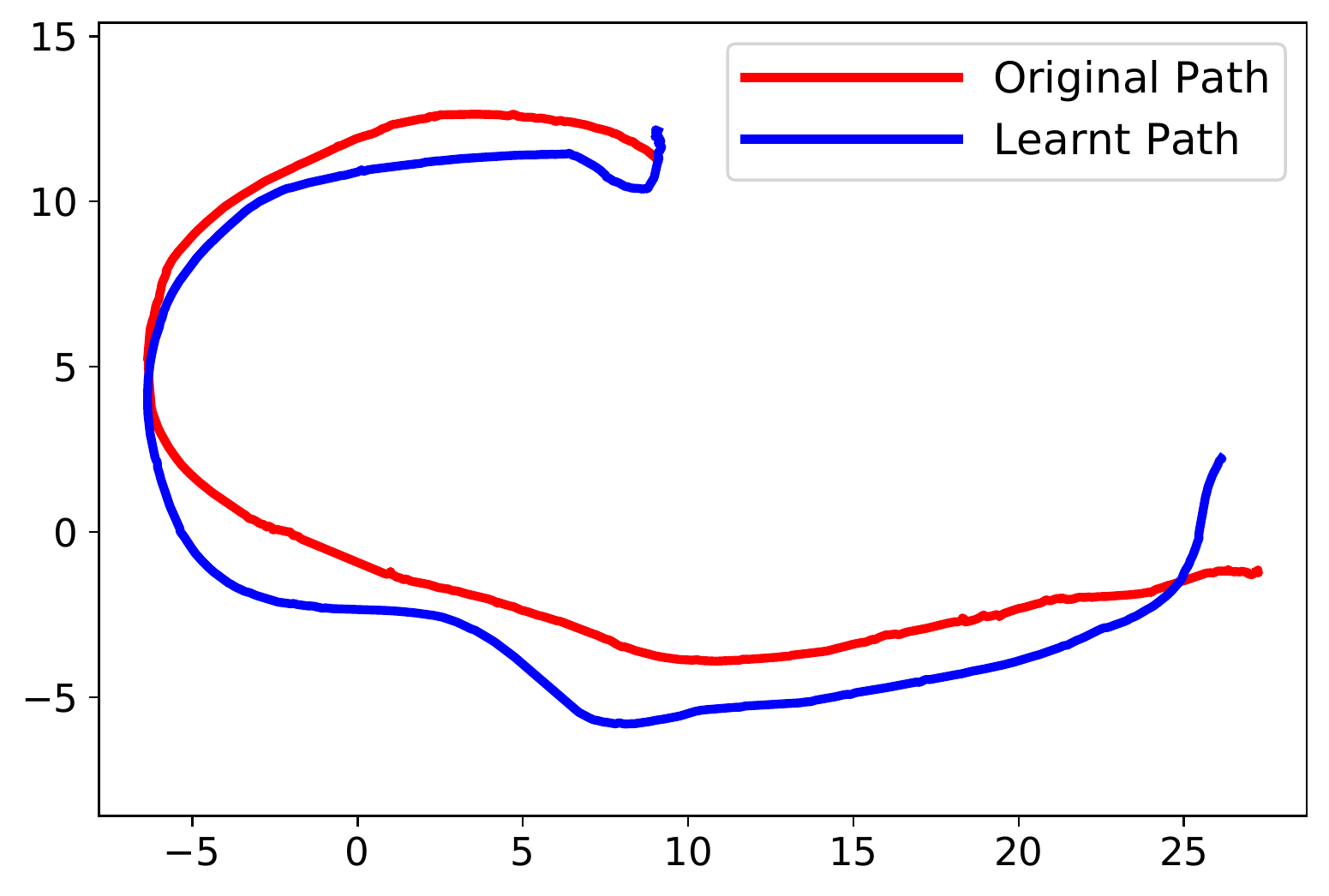}} 
    \subfigure[Speech Commands]{\includegraphics[width=0.48\columnwidth]{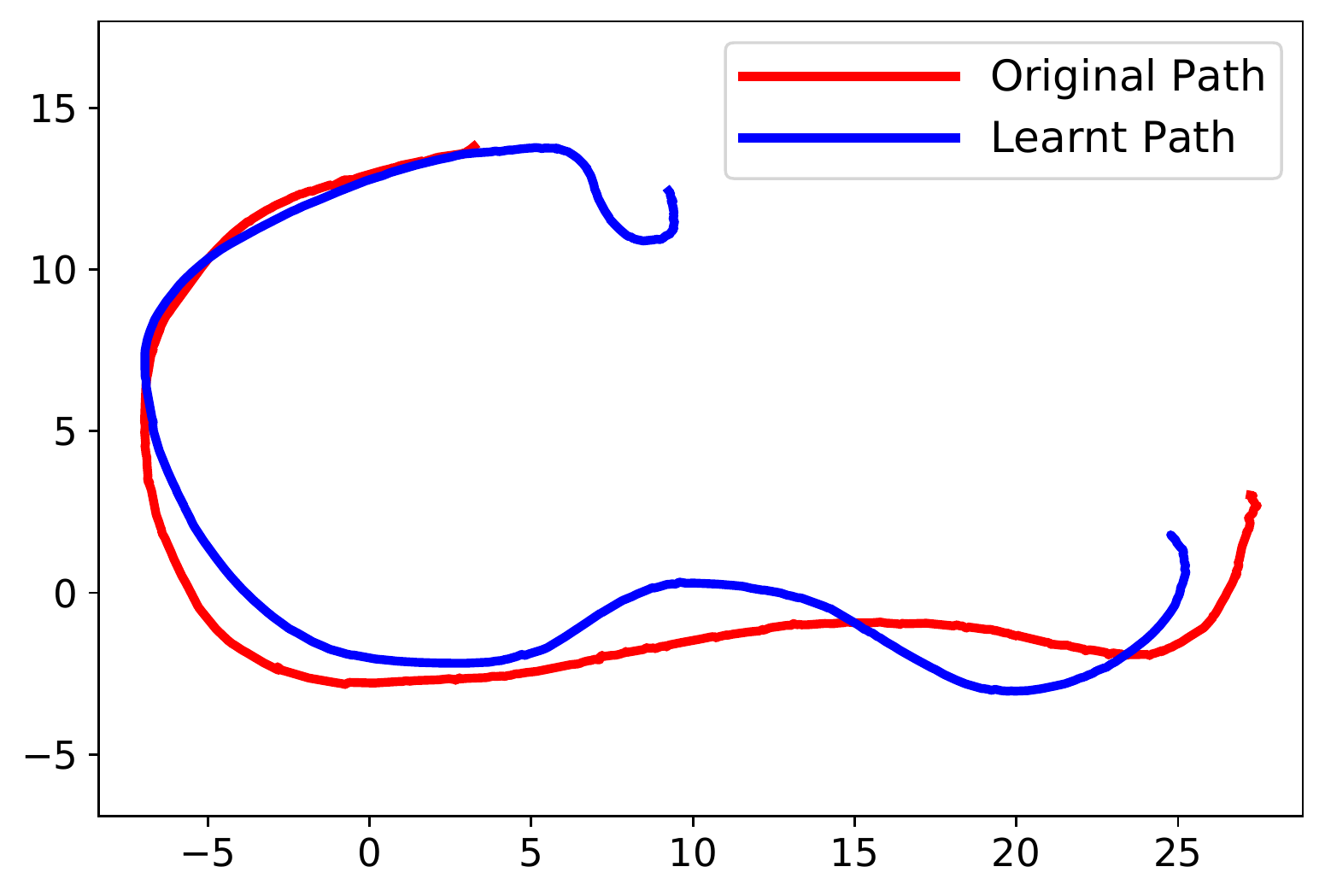}}
    \subfigure[Speech Commands]{\includegraphics[width=0.48\columnwidth]{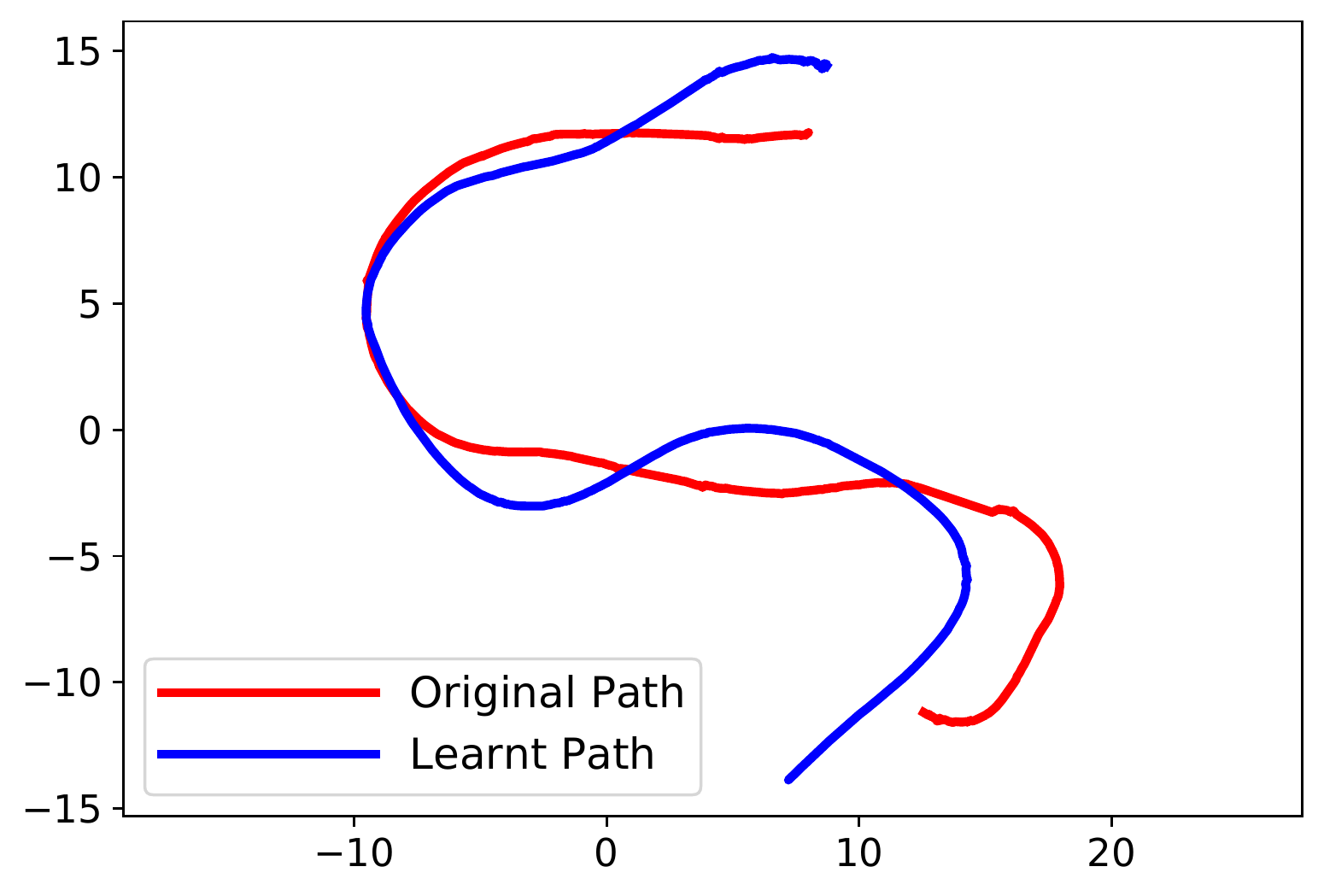}} 
    \caption{Examples of the original $X$ and learnt path $Y$ in Character Trajectory, PhysioNet Sepsis and Speech Commands . We use UMAP~\cite{NBC2020} to project each time-series onto a 2-dim space for visualization. The learnt path is similar to the original path but with some modifications.} 
    \label{fig:append_umap2}
    
\end{figure*}

We also visualize the original path $X$ and the learnt path $Y$ in various datasets in Fig.~\ref{fig:append_umap2}.

\section{Sensitivity to $\alpha, \beta$}\label{sec:sensitivity}
\begin{figure*}[ht]
    \centering
   
    \subfigure[Speech Commands (larger accuracy values are preferred.)]{\includegraphics[width=0.65\columnwidth]{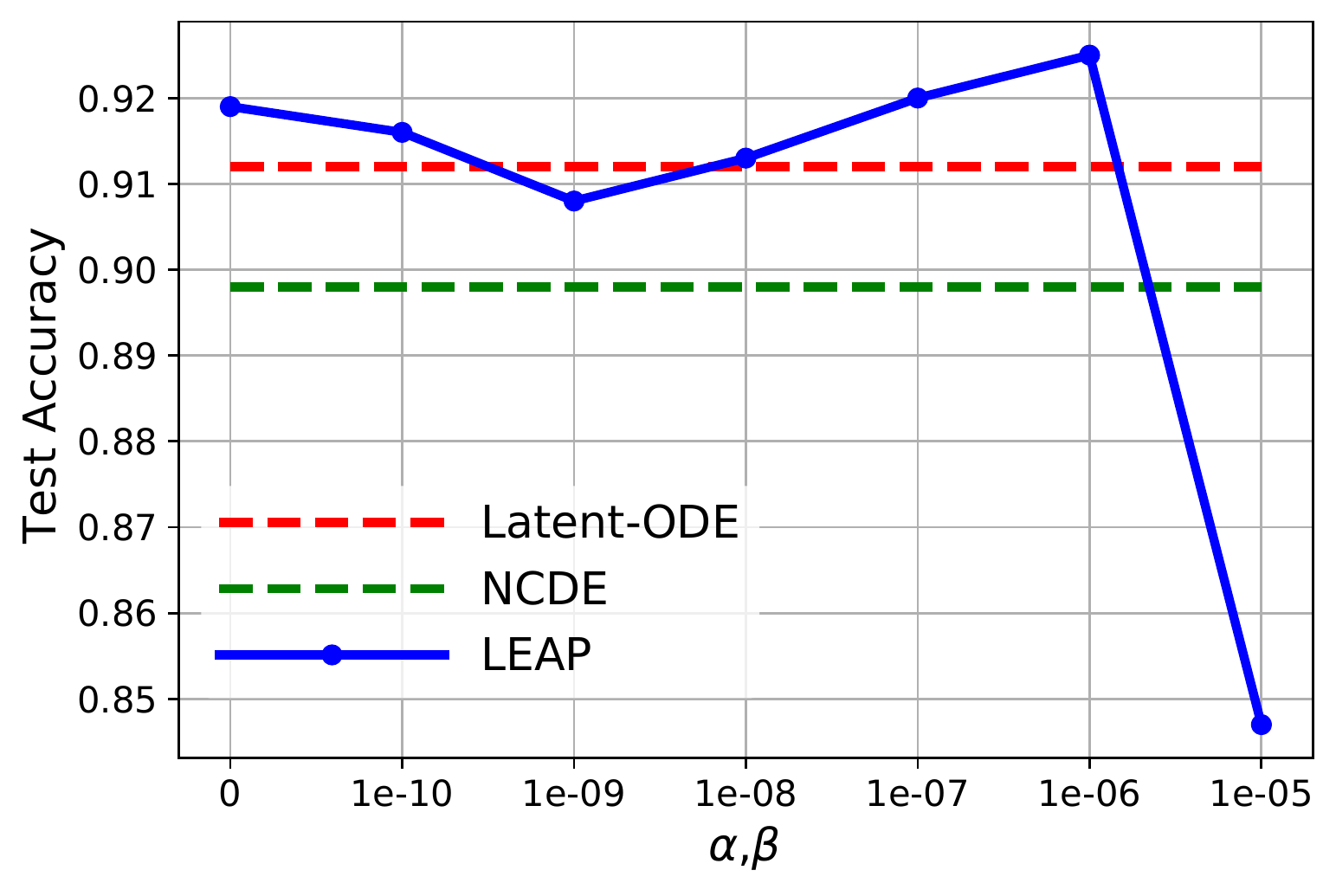}}
    \subfigure[PhysioNet Sepsis (larger AUROC values are preferred.)]{\includegraphics[width=0.65\columnwidth]{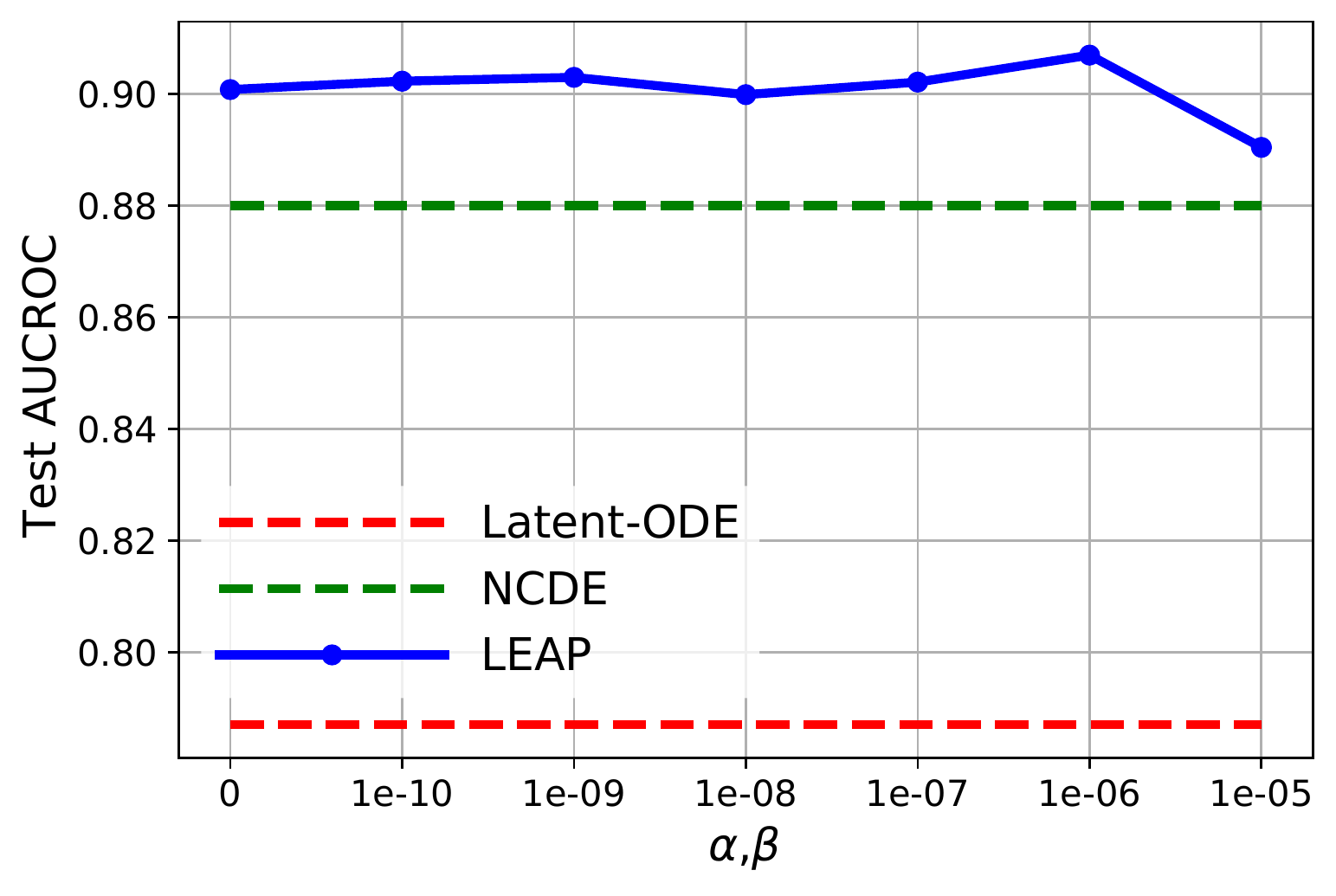}}
    \subfigure[MuJoCo (smaller error values are preferred.)]{\includegraphics[width=0.65\columnwidth]{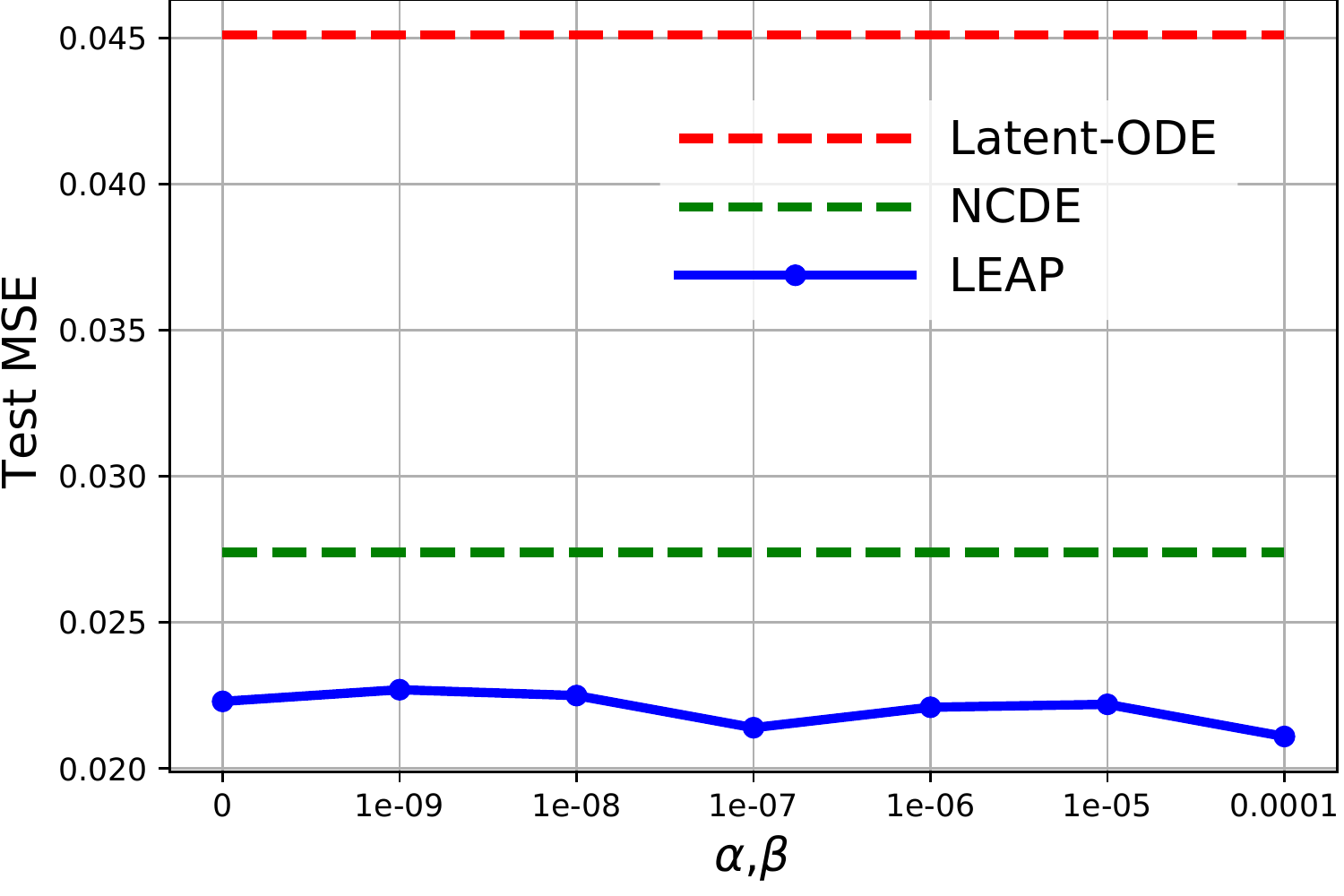}}
    \caption{Sensitivity to $\alpha, \beta$ in terms of their various values} 
    
    \label{fig:append_alpha_beta}
    
\end{figure*}

\begin{figure*}[t]
    \centering
    \subfigure[Character Trajectories]{\includegraphics[width=0.49\columnwidth]{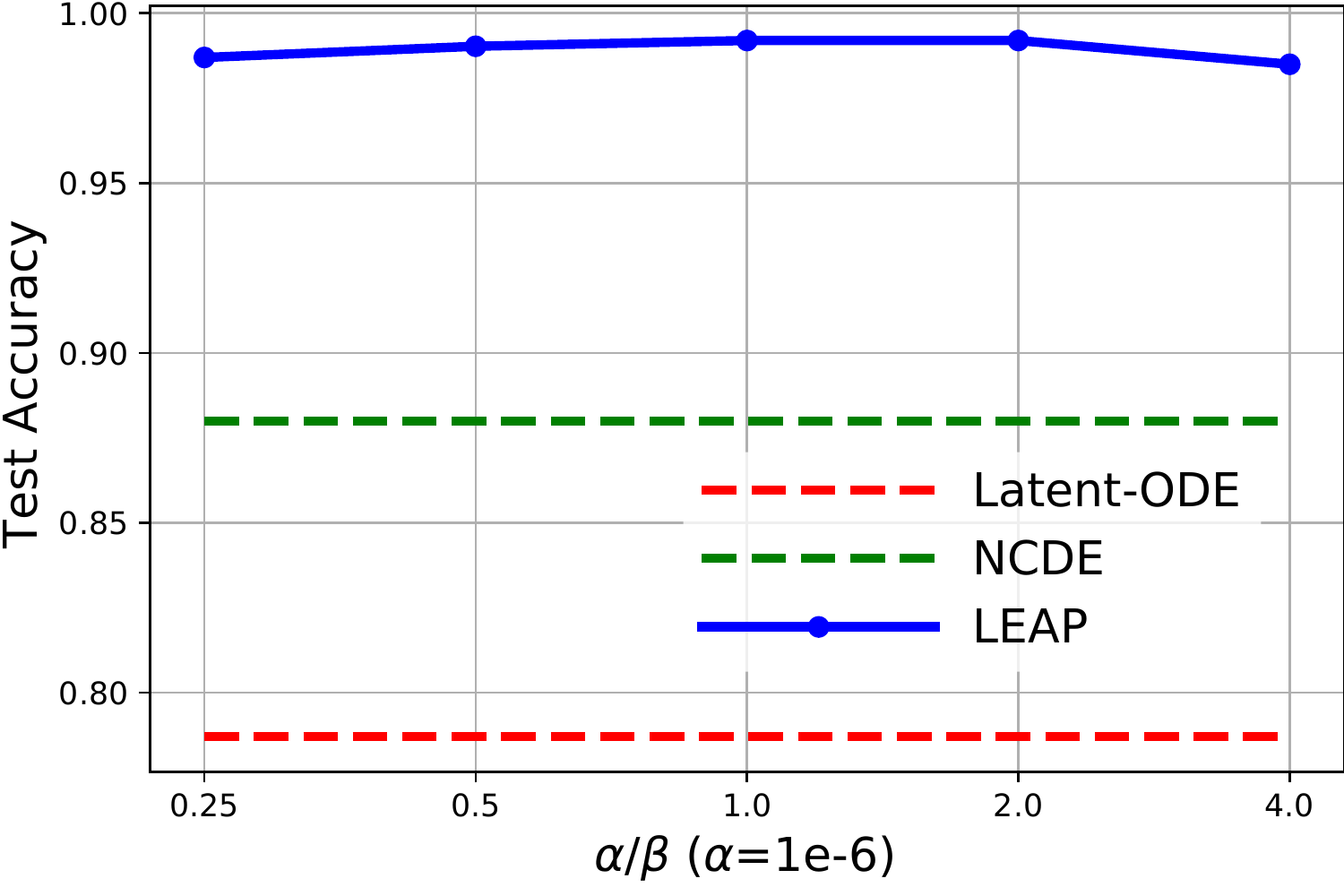}}
    \subfigure[PhysioNet Sepsis]{\includegraphics[width=0.49\columnwidth]{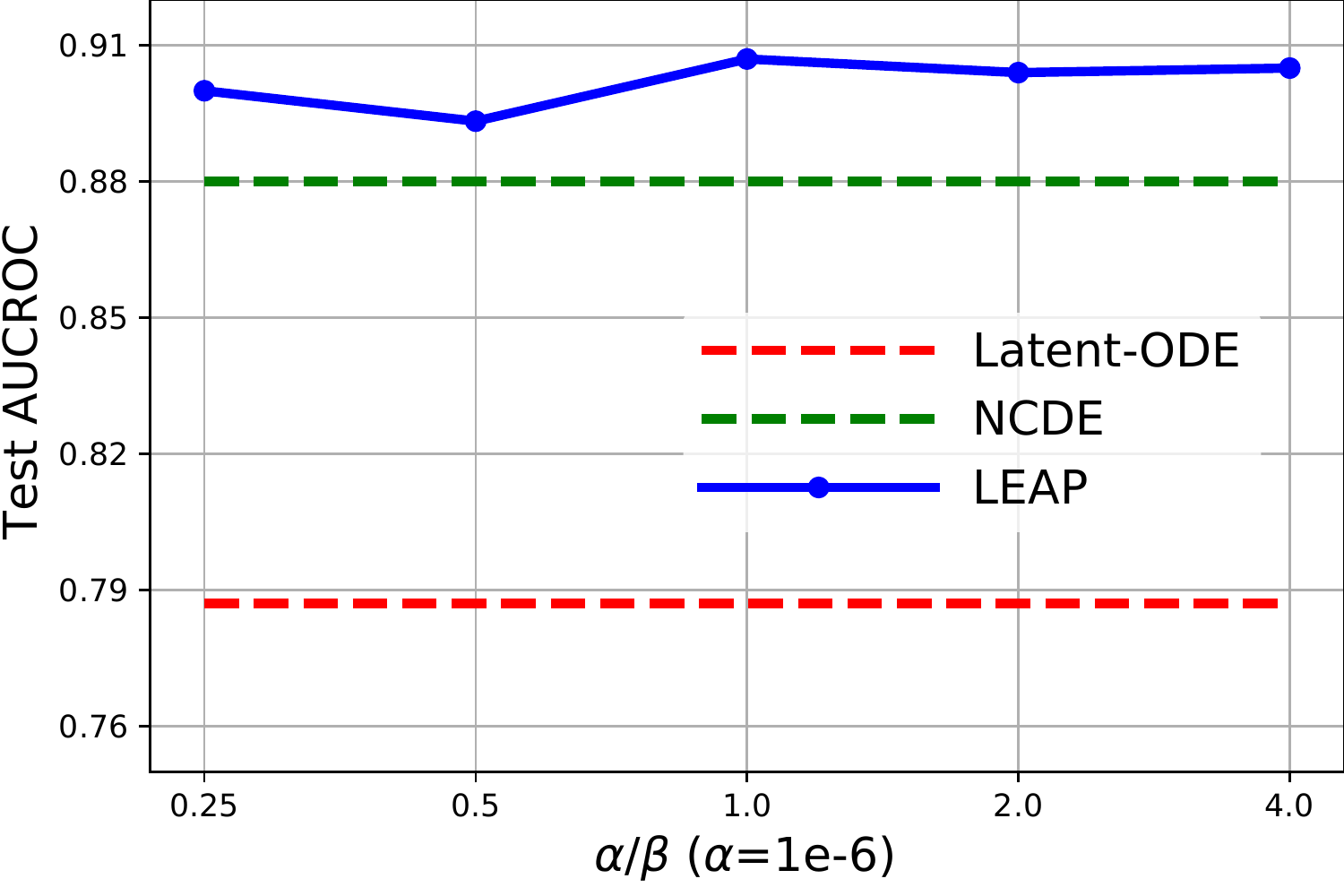}}
    \subfigure[Speech Commands]{\includegraphics[width=0.49\columnwidth]{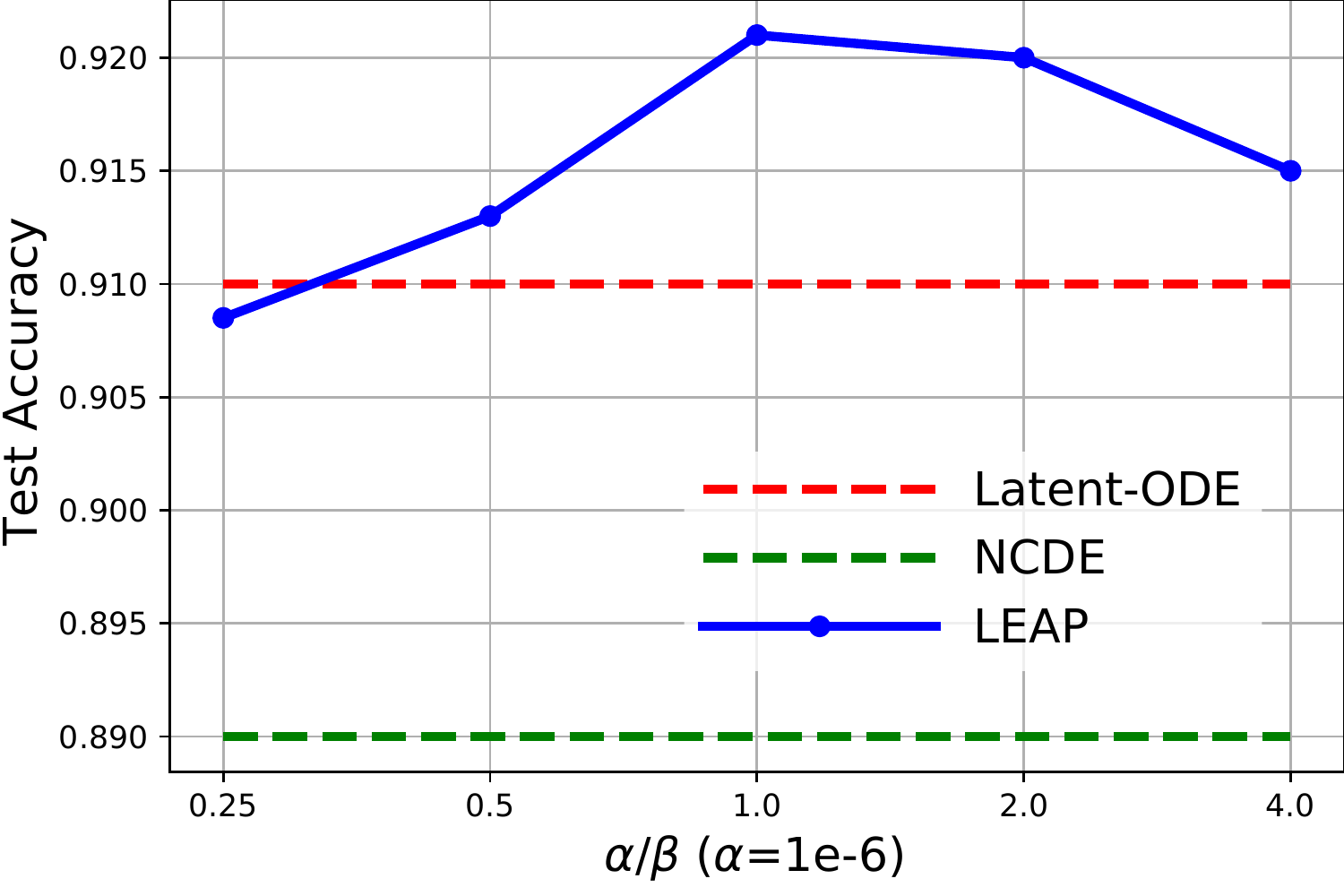}}
    \subfigure[MuJoCo]{\includegraphics[width=0.49\columnwidth]{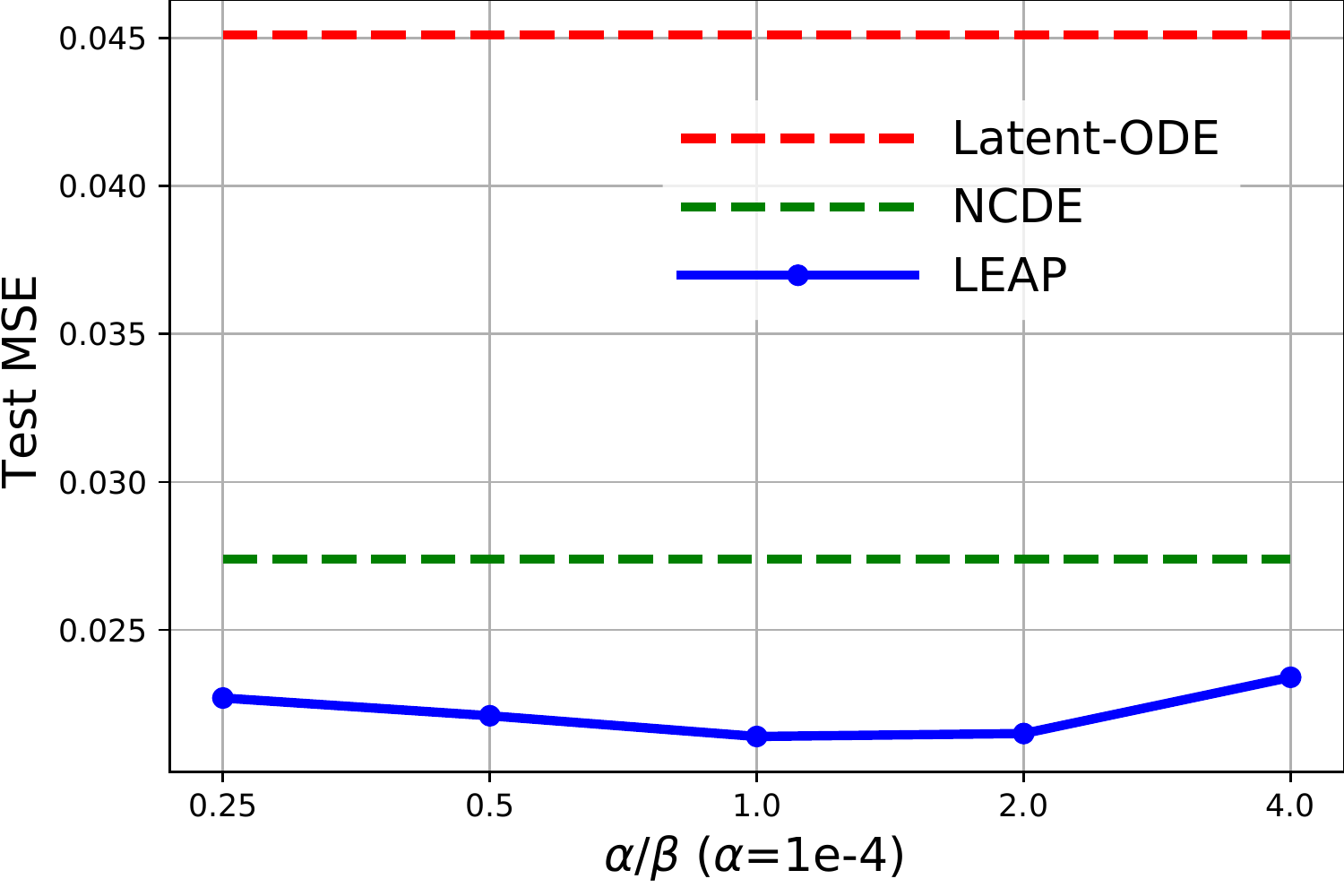}}
    \caption{Sensitivity to $\alpha, \beta$ in terms of their various ratios}
    \label{fig:ratio_diff}
\end{figure*}
Fig.~\ref{fig:append_alpha_beta} shows the sensitivity curve w.r.t. $\alpha, \beta$, while setting $\alpha$ and $\beta$ to same value, i.e. the ratio of $\alpha$ to $\beta$ is 1, as in our original setting. Fig.~\ref{fig:append_alpha_beta} (a), the model accuracy fluctuates, but still shows the best score in several settings, compared to baselines. The model AUROC and MSE of Fig.~\ref{fig:append_alpha_beta} (b), and (c) are robust in all settings and always show better performances than those of baselines. On the contrary, in the case of  From those results, we found that moderate $\alpha, \beta$ settings (i.e., not too small and not too large) are needed to achieve the best accuracy or MSE for all datasets. 

We also report the additional results in Fig.~\ref{fig:ratio_diff} for various ratios of ${\alpha}$ to ${\beta}$. As shown, a good trade-off between $\alpha$ and $\beta$ is required. In general, their ratio should be around 1.0 in our experiments to see reasonable results. If their settings are biased toward one of them, the overall accuracy gets worse. However, LEAP still shows the highest performance in all ratio settings but one case. 

\section{Reproducibility Checklist}

\paragraph{Randomness}
We test our experiments as following random seeds, and also conduct baselines with same random seeds. 
\begin{enumerate}
    \item Character Trajectories : 112,198,234,307,425
    \item PhysioNet Sepsis  : 112,176,234,929,8888
    \item Speech Commands : 112,118,198,234,824
    \item MuJoCo : 118,176,234,345,824
\end{enumerate}
\paragraph{Criteria used to select final parameter settings}
The final parameters were chosen based on reasonable memory usage (not to much exceed the baseline memory usage) and good performance.
\end{appendices}
\end{document}